\documentclass[journal]{IEEEtran}
\usepackage{cite}
\usepackage{pdfpages}
\usepackage{graphicx}
\usepackage{subfloat}
\usepackage{caption}
\usepackage{subcaption}
\usepackage{url}
\usepackage{makecell}
\usepackage{float}
\usepackage{amsmath}
\usepackage{mathtools}
\usepackage{amsfonts}
\usepackage{enumitem}
\usepackage{amssymb}
\usepackage{algorithm2e,algpseudocode}

\DeclareMathOperator*{\argmax}{arg\,max}

\ifCLASSINFOpdf
\else
\fi
\SetKwProg{Fn}{Function}{}{}
\begin{document}
%
\title{Deep Q-Learning with Q-Matrix Transfer Learning for Novel Fire Evacuation Environment}

\author{\IEEEauthorblockN{Jivitesh Sharma\IEEEauthorrefmark{1},
Per-Arne Andersen\IEEEauthorrefmark{2}, Ole-Christoffer Granmo\IEEEauthorrefmark{3} and
Morten Goodwin\IEEEauthorrefmark{4}}\\
\small{\IEEEauthorblockA{Centre for Artificial Intelligence Research \\ Department of Information and Communication Technology\\
University of Agder, Norway\\
\IEEEauthorrefmark{1}jivitesh.sharma@uia.no,
\IEEEauthorrefmark{2}per@sysx.no,
\IEEEauthorrefmark{3}ole.granmo@uia.no,
\IEEEauthorrefmark{4}morten.goodwin@uia.no}}
}


\maketitle

\begin{abstract}
Deep Reinforcement Learning is achieving significant success in various applications like control, robotics, games, resource management, and scheduling. However, the important problem of emergency evacuation, which clearly could benefit from reinforcement learning, has been largely unaddressed. Indeed, emergency evacuation is a complex task which is difficult to solve with reinforcement learning. An emergency situation is highly dynamic, with a lot of changing variables and complex constraints that make it challenging to solve. Also, there is no standard benchmark environment available that can be used to train Reinforcement Learning agents for evacuation. A realistic environment can be complex to design. In this paper, we propose the first fire evacuation environment to train reinforcement learning agents for evacuation planning. The environment is modelled as a graph capturing the building structure. It consists of realistic features like fire spread, uncertainty and bottlenecks. We have implemented the environment in the OpenAI gym format, to facilitate future research. We also propose a new reinforcement learning approach that entails pretraining the network weights of a DQN based agent (DQN/Double-DQN/Dueling-DQN) to incorporate information on the shortest path to the exit. We achieved this by using tabular Q-learning to learn the shortest path on the building model's graph. This information is transferred to the network by deliberately overfitting it on the Q-matrix. Then, the pretrained DQN model is trained on the fire evacuation environment to generate the optimal evacuation path under time varying conditions due to fire spread, bottlenecks and uncertainty. We perform comparisons of the proposed approach with state-of-the-art reinforcement learning algorithms like DQN, DDQN, Dueling-DQN, PPO, VPG, SARSA, A2C and ACKTR. The results show that our method is able to outperform state-of-the-art models by a huge margin including the original DQN based models. Finally, we test our model on a large and complex real building consisting of $91$ rooms, with the possibility to move to any other room, hence giving $8281$ actions. In order to reduce the action space, we propose a strategy that involves one step simulation. That is, an action importance vector is added to the final output of the pretrained DQN and acts like an attention mechanism. Using this strategy, the action space is reduced by $90.1\%$. In this manner, we are able to deal with large action spaces. Hence, our model achieves near optimal performance on the real world emergency environment.
\end{abstract}


\IEEEpeerreviewmaketitle

\section{Introduction}
\IEEEPARstart{R}{einforcement} Learning (RL) has been a subject of extensive research and applications in various real world domains such as Robotics, Games, Industrial Automation and Control, System Optimization, Quality Control and Maintenance. But, some extremely important areas, where Reinforcement Learning could be immensely vital, have not received adequate attention from researchers. We turn our attention to the major problem of evacuation in case of fire emergencies.\\
Fire related disasters are the most common type of Emergency situation. They require thorough analysis of the situation for quick and precise response. Even though this critical application hasn't received adequate attention from AI researchers, there have been some noteworthy contributions. One such paper, focusing on assisting decision making for fire brigades, is described in \cite{BRIGADE}. Here, the the RoboCup Rescue simulation is used as a fire simulation environment \cite{ROBO}. A SARSA Agent \cite{SARSA} is used with a new learning strategy called Lesson-by-Lesson learning, similar to curriculum learning. Results show that the RL agent is able to perform admirably in the simulator. However, the simulator lacks realistic features like bottlenecks, fire spread and has a grid structure which is too simplistic to model realistic environments. Also, the approach seems unstable and needs information about the state which isn't readily available in real life scenarios.\\
In \cite{AGENT}, multiple coordinated agents are used for forest fire fighting. The paper uses a software platform called Pyrosim which is used to create dynamic forest fire situations. The simulator is mostly used for terrain modeling and a coordinated multiple agent system is used to extinguish fire and not for evacuation. \\
The evacuation approach described in \cite{INDOOR} is similar to the problem we try to solve in this paper. In \cite{INDOOR}, a fading memory mechanism is proposed with the intuition that in dynamic environments less trust should be put on older knowledge for decision making. But arguably, this could be achieved more efficiently by the '$\gamma$' parameter in Q-learning along with prioritized experience replay. Also, the graph based environment used in \cite{INDOOR} lacks many key features like fire spread, people in rooms, bottlenecks etc. \\
The most significant work done on building evacuation using RL is reported in \cite{THESIS}. The evacuation environment is grid based with multiple rooms and fire. The fire spread is modelled accurately and uncertainty taken into account. The multi-agent Q-learning model is shown to work in large spaces as well. Further, the paper demonstrates a simple environment and strategy for evacuation. However, the approach proposed in \cite{THESIS} lacks key features like bottlenecks and actual people in rooms. The grid based environment isn't able to capture details of the building model like room locations and paths connecting rooms. \\
Some interesting research on evacuation planning take a completely different approach by simulating and modelling human and crowd behaviour under evacuation \cite{SAFEGRESS,EGRESS,BDI,BDI2}. Our work on evacuation planning is not based on human behaviour modelling or the BDI (Belief-Desire-Intention) framework for emergency scenarios. These methods are beyond the scope of this paper and not discussed here.\\
\subsubsection*{\textbf{Proposed Environment}}
There are many reinforcement learning libraries that contain simulations and game environments to train reinforcement learning based agents \cite{GYM,UNITY,SC2,PYG,SIM}. However, currently no realistic learning environment for emergency evacuation has been reported.\\
In our paper, we build the first realistic fire evacuation environment specifically designed to train reinforcement learning agents for evacuating people in the safest manner in the least number of time-steps possible. The environment has the same structure as OpenAI gym environments, so it can be used easily in the same manner.\\
The proposed fire evacuation environment is graph based, which requires complex decision making such as routing, scheduling and dealing with bottlenecks, crowd behaviour uncertainty and fire spread. This problem falls in the domain of discrete control. The evacuation is performed inside a building model, which is represented as a graph. The agent needs to evacuate all persons in all rooms through any available exits using the shortest path in the least number of time-steps, while avoiding any perilous situations like the fire, bottlenecks and other hazardous situations.\\
Some previous research papers focus on modelling fire spread and prediction, mostly using cellular automata \cite{CELLULAR} and other novel AI techniques \cite{CELLULAR2,CONTROL,AIFOREST}. An effective and innovative way of modelling fire spread is to use spatial reinforcement learning, as proposed in \cite{SPREAD}. However, our way of simulating fire spread is far less complex and leverages rewarding system of the RL framework. In our proposed environment, we simply use an exponential decay reward function to model the fire spread and direction. To keep in tune with the RL framework, the feedback from the environment sent back to the agent should convey enough information. So, we design the reward function in such a manner that the agent can learn about the fire spread and take measures accordingly.\\

\subsubsection*{\textbf{Proposed Method}}
Since this environment poses a high level of difficulty, we argue that incorporating the shortest path information (shortest path from each room to the nearest exit) in the DQN model(s) by transfer learning and pretraining the DQN neural network function approximator is necessary. \\
Transfer learning has been used extensively in computer vision tasks for many years, recently vastly expanded for many computer vision problems in \cite{TASK}. Lately, it has been utilized in Natural Language models \cite{LM,BERT}. In reinforcement learning, pretrained models have started to appear as well \cite{WORLD,TTQL}. In fact, we use the convergence analysis of \cite{TTQL}, which provides a general theoretical perspective of task transfer learning, to prove that our method guarantees convergence.\\
In this paper, we present a new class of pretrained DQN models called Q-matrix Pretrained Deep Q-Networks (QMP-DQN). We employ Q-learning to learn a Q-matrix representing the shortest paths from each room to the exit. We perform multiple random episodic starts and $\epsilon$-greedy exploration of the building model graph environment. Q-learning is applied on a pretraining instance of the environment that consists of only the building model graph. Then, we transfer the Q-matrix to a DQN model, by pretraining the DQN to reproduce the Q-matrix. Finally, we train the pretrained DQN agent on the complete fire evacuation task. We compare our proposed pretrained DQN models (QMP-DQN) against regular DQN models and show that pretraining for our fire evacuation environment is necessary. We also compare our QMP-DQN models with state-of-the-art Reinforcement Learning algorithms and show that off-policy Q-learning techniques perform better than other policy based methods as well as actor-critic models.\\
Finally, in Section 5, we show that our method can perform optimal evacuation planning on a large and complex real world building model by dealing with the large discrete action space in a new and simple way by using an attention based mechanism.\\
\subsubsection*{\textbf{Contributions}}
This paper contributes to the field of reinforcement learning, emergency evacuation and management in the following manner:
\begin{enumerate}
  \item{We propose the first reinforcement learning based fire evacuation environment with OpenAI Gym structure.}
  \item{We build a graph based environment to accurately model the building structure, which is more efficient than a maze structure.}
  \item{The environment can consist of a large discrete action space with $n^2$ number of actions (for all possibilities), where $n$ is the number of rooms in the building. That is, the action space size increases exponentially with respect to the rooms.}
  \item{Our proposed environment contains realistic features such as multiple fires and dynamic fire spread which is modelled by the exponential decay reward function.}
  \item{We further improve the realism of our environment by restricting the number of people allowed in each room to model over-crowded hazardous situations.}
  \item{We also include uncertainty about action performed in the environment to model uncertain crowd behaviour, which also acts as a method of regularization.}
  \item{We use the Q-matrix to transfer learned knowledge of the shortest path by pretraining a DQN agent to reproduce the Q-matrix.}
  \item{We also introduce a small amount of noise in the Q-matrix, to avoid stagnation of the DQN agent in a local optimum.}
  \item{We perform exhaustive comparisons with other state-of-the-art reinforcement learning algorithms like DQN, DDQN, Dueling DQN, VPG, PPO, SARSA, A2C and ACKTR.}
  \item{We test our model on a large and complex real world scenario, which is the University of Agder Building, which consists of $91$ nodes, and $8281$ actions.}
  \item{We propose a new and simple way to deal with large discrete action spaces in our proposed environment, by employing an attention mechanism based technique.}
\end{enumerate}

The rest of the paper is organized as follows: Section 2 summarizes the RL concepts used in this paper. Section 3 gives a detailed explanation of the proposed Fire Emergency Evacuation System with each module of the system described in subsequent sub-sections. Section 4 reports our exhaustive experimental results. Section 5 presents the real world application of our model in a large and complex environment, and finally Section 6 concludes the paper.

\section{Preliminaries}
Reinforcement Learning is a sub-field of Machine Learning which deals with learning to make appropriate decisions and take actions to achieve a goal. A Reinforcement Learning agent learns from direct interactions with an environment without requiring explicit supervision or a complete model of the environment. The agent interacts with the environment by performing actions. It receives feedback for it's actions in terms of reward (or penalty) from the environment and observes changes in the environment as a result of the actions it performs. These observations are called states of the environment and the agent interacts with the environment at discrete time intervals $t$ by performing an action $a_t$ in a state of the environment $s_t$, it transitions to a new state $s_{t+1}$ (change in the environment) while receiving a reward $r_t$, with probability $P(s_{t+1}|s_t,a_t)$. The main aim of the agent is to maximize the cumulative reward over time through it's choice of actions. A pictorial representation of the RL framework is shown in Fig. 1.\\
In the subsequent subsections, a brief presentation of the concepts and methods used in this paper are explained.

\begin{figure}[H]
  \includegraphics[width=\linewidth]{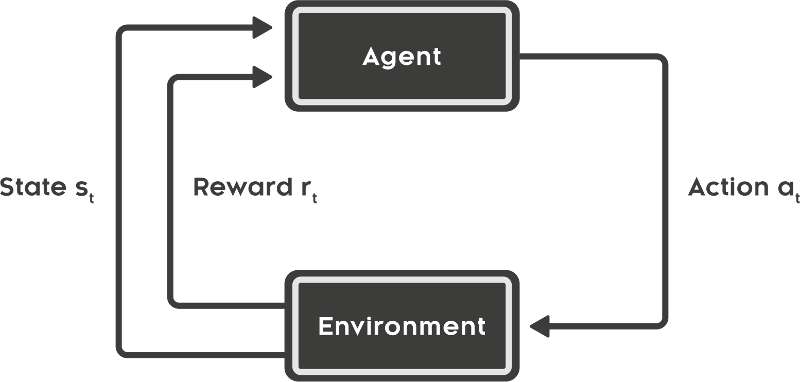}
  \label{fig:sfig1}
  \caption{Reinforcement Learning Framework (the figure is taken from \cite{SUTTON})}
\end{figure}

\subsection{\textbf{Markov Decision Process}}
The Reinforcement learning framework is formalised by Markov Decision Processes (MDP) which are used to define the interaction between a learning agent and its environment in terms of states, actions, and rewards \cite{MDP}. An MDP consists of a tuple of $\langle S,A,P,R \rangle$ \cite{SUTTON}, where $S$ is the state space, $A$ is the action space, $P$ is the transition probability from one state to the next, $P: S\times A\times S\longmapsto [0,1]$ and $R$ is the reward function, $R: S\times A\longmapsto \mathbb{R}$.\\
When state space $S$, action space $A$ and rewards $R$ consist of finite number of elements, $s_{t+1}$ and $r_{t+1}$ have well-defined discrete probability distributions which depend only on the present state and action (Markov Property). This is represented as $p(s_{t+1},r_{t+1}|s_t,a_t)$, where $p$ determines the dynamics of the Markov Decision Process and where:
\begin{equation}\label{1}
  \sum_{s_{t+1}\in S}\sum_{r\in R} p(s_{t+1},r_{t+1}|s_t,a_t) = 1, \forall s_t\in S, a_t\in A
\end{equation}
$p$ contains all the information about the MDP, so we can compute important aspects about the environment from $p$, like state transition probability and expected rewards for state-action pairs \cite{SUTTON}:
\begin{equation}\label{2}
  P(s_{t+1}|s_t,a_t) = \sum_{r\in R} p(s_{t+1},r_{t+1}|s_t,a_t)
\end{equation}
\begin{equation}\label{3}
  r(s_t,a_t) = \mathbb{E}[r_t|s_t,a_t] = \sum_{r\in R} r\sum_{s_{t+1}\in S} p(s_{t+1},r_{t+1}|s_t,a_t)
\end{equation}
The equation 3, gives the immediate reward we expect to get when performing action $a_t$ from state $s_t$. The agent tries to select actions that maximize the sum of rewards it expects to achieve, as time goes to infinity. But, in a dynamic and/or continuous Markov Decision Process, the notion of discounted rewards is used \cite{SUTTON}:
\begin{equation}\label{4}
  G_t = \sum_{k=0}^{\infty} \gamma^kr_{t+k+1}
\end{equation}
where, $\gamma$ is the discount factor and is in the range $[0,1]$. If $\gamma$ is near $0$, then the agent puts emphasis on rewards received in the near future and if $\gamma$ is near $1$, then the agent also cares about rewards in the distant future.\\
In order to maximize $G_t$, the agent picks an action $a_t$ when in a state $s_t$ according to a policy function $\pi(s_t)$. A policy function is a probabilistic mapping from the state space to the action space, $S\rightarrow A$. The policy function outputs probabilities for taking each action in give state, so it can also be denoted as $\pi(a_t|s_t)$.

\subsection{\textbf{Q-Learning}}
Most of the Reinforcement Learning algorithms (value based) try to estimate the value function which gives an estimate of how good a state is for the agent to reside in. This is estimated according to the expected reward of a state under a policy and is denoted as $v_\pi(s)$:
\begin{equation}\label{5}
  v_\pi(s) = \mathbb{E}_\pi[G_t|s_t]
\end{equation}
Q-learning is a value based Reinforcement Learning algorithm that tries to maximize the $q$ function \cite{Q}. The $q$ function is a state-action value function and is denoted by $Q(s_t,a_t)$. It tries to maximize the expected reward give a state and action performed on that state:
\begin{equation}\label{6}
  Q(s_t,a_t) = \mathbb{E}[G_t|s_t,a_t]
\end{equation}
According the Bellman Optimality equation \cite{SUTTON}, the optimal $q$ function can be obtained by:
\begin{align}
  Q^\ast(s_t,a_t)&=\mathbb{E}[r_{t+1}+\gamma v^\ast(s_{t+1})|s_t,a_t] \nonumber \\
  &=\sum_{s_{t+1},r_t}p(s_{t+1},r_t|s_t,a_t)[r_t+\gamma v^\ast(s)]
\end{align}
where, $v^\ast(s_{t+1}) = \max_{a_{t+1}}Q^\ast(s_{t+1},a_{t+1})$. And, $a^\ast$ is the optimal action which results in maximum reward, the optimal policy is formed as $\argmax_{a_t}\pi^\ast(a_t|s_t) = a^\ast$. This method was proposed in \cite{Q} which is tabular style Q-learning. The update rule for each time step of Q-learning is as follows:
\begin{multline}\label{8}
    Q_{t+1}(s_t,a_t) = Q_t(s_t,a_t)+\eta[r_t+\gamma\max_{a_t}Q_t(s_{t+1},a_{t+1})\\-Q_t(s_t,a_t)]
 \end{multline}
Q-learning is an incremental dynamic programming algorithm that determines the optimal policy in a step-by-step manner. At each step $t$, the agent performs the following operations:
\begin{itemize}
  \item{Observes current state $s_t$.}
  \item{Selects and performs an action $a_t$.}
  \item{Observes the next state $s_{t+1}$.}
  \item{Receives the reward $r_t$.}
  \item{Updates the q-values $Q_t(s_t,a_t)$ using equation 8.}
\end{itemize}
The $q$ value function converges to the optimal value $Q_{t+1}(s_t,a_t)\rightarrow Q^\ast(s_t,a_t)$ as $t\rightarrow\infty$. Detailed convergence proof and analysis can be found in \cite{Q}.\\
This tabular Q-learning method is used in our proposed approach to generate a Q-matrix for the shortest path to the exit based on the building model. In order to incorporate the shortest path information, this Q-matrix is used to pretrain the DQN models.

\subsection{\textbf{Deep Q Network}}
The tabular Q-learning approach works well for small environments, but becomes infeasible for complex environments with large multidimensional discrete or continuous state-action spaces. To deal with this problem, a parameterized version of the $q$ function is used for approximation $Q(s_t,a_t;\theta)\approx Q^\ast(s_t,a_t)$. This way of function approximation was first proposed in \cite{FIRST}. \\
Deep Neural Networks (DNNs) have become the predominant method for approximating complex intractable functions. They have become the defacto method for various applications such as image processing and classification \cite{IMAGENET,CNNOriginal,VGG16,RESNET,LeNet-5,XCEPTION,ALEXNET}, speech recognition \cite{LAS,GOOGLESPEECH,E2ESRRNN,GHINT,ALEX,LVCSR,CONTEXT}, and natural language processing \cite{NLP1,NLP2,NLP3,NLP4,NLP5}. DNNs have also been applied to reinforcement learning problems successfully by achieving noteworthy performance \cite{DNNRL1,DNNRL2}. \\
The most noteworthy research in integrating deep neural networks and Q-learning in an end-to-end reinforcement learning fashion is the Deep Q-Networks (DQNs) \cite{DQN1,DQN2}. To deal with the curse of dimensionality, a neural network is used to approximate the parameterised Q-function $Q(s_t,a_t;\theta)$. The neural network takes a state as input and approximates Q-values for each action based on the input state. The parameters are updated and the Q-function is refined in every iteration through an appropriate optimizer like Stochastic Gradient Descent \cite{SGD}, RMSProp \cite{RMS}, Adagrad \cite{ADAGRAD}, Adam \cite{ADAM} etc. The neural network outputs q-values for each action for the input state and the action with the highest q-value is selected (There is another DQN architecture, which is less frequently used, that takes in the state and action as input and returns it's q-value as output).\\
The DQN can be trained by optimizing the following loss function:
\begin{equation}\label{9}
  L_i(\theta_i) = \mathbb{E}[(r_t+\gamma\max_{a_{t+1}}Q(s_{t+1},a_{t+1};\theta_{i-1})-Q(s_{t},a_{t};\theta_{i}))^2]
\end{equation}
where, $\gamma$ is the discount factor, $\theta_i$ and $\theta_{i-1}$ are the Q-network parameters at iteration $i$ and $i-1$ respectively. In order to train the Q-network, we require a target to calculate loss and optimize parameters. The target q-values are obtained by holding the parameters $\theta_{i-1}$ fixed from the previous iteration.
\begin{equation}\label{10}
  y = r_t+\gamma\max_{a_{t+1}}Q(s_{t+1},a_{t+1};\theta_{i-1})
\end{equation}
where, $y$ is the target for the next iteration to refine the Q-network. Unlike supervised learning where the optimal target values are known and fixed prior to learning, in DQN the approximate target values $y$, which depend on network parameters, are used to train the Q-network. The loss function can be rewritten as:
\begin{equation}\label{11}
  L_i(\theta_i) = \mathbb{E}[(y-Q(s_{t},a_{t};\theta_{i}))^2]
\end{equation}
The process of optimizing the loss function $L_i(\theta_i)$ at the $i^{th}$ iteration by holding the parameters from the previous iteration $\theta_{i-1}$ fixed, to get target values, results in a sequence of well-defined optimization time-steps. By differentiating the loss function in equation 11, we get the following gradient:
\begin{equation}\label{12}
  \nabla_{\theta_i}L_i(\theta_i) =\mathbb{E}[(y-Q(s_{t},a_{t};\theta_{i}))\nabla_{\theta_i}Q(s_{t},a_{t};\theta_{i})]
\end{equation}\label{12}
Instead of computing the full expectation of the above gradient, we optimize the loss function using an appropriate optimizer (in this paper we use the Adam optimizer \cite{ADAM}). The DQN is a model-free algorithm since it directly solves tasks without explicitly estimating the environment dynamics. Also, DQN is an off-policy method as it learns a greedy policy $a=\argmax_{a_{t+1}}Q(s,a_{t+1};\theta)$, while following an $\epsilon$-greedy policy for sufficient exploration of the state space.
One of the drawbacks of using a nonlinear function approximator like neural network is that it tends to diverge and is quite unstable for reinforcement learning. The problem of instability arises mostly due to: correlations between subsequent observations and that small changes in q-values can significantly change the policy and the correlations between q-values and target values.\\
The most well-known and simple technique to alleviate the problem of instability is the experience replay \cite{RLNN}. At each time-step, a tuple consisting of the agent's experience $E_t=(s_t,a_t,r_t,s_{t+1})$ is stored in a replay memory over many episodes. A minibatch of these tuples is randomly drawn from the replay memory to update the DQN parameters. This ensures that the network isn't trained on a sequence of observations (avoiding strong correlations between samples and reducing variance between updates) and it increases sample efficiency. This technique greatly increases stability of DQN.

\subsection{\textbf{Double DQN}}
Q-learning and DQN are capable of achieving performance beyond the human level on many occasions. However, in some cases Q-learning performs poorly and so does its deep neural network counterpart DQN. The main reason behind such poor performance is that Q-learning tends to overestimate action values. These overestimations are caused due to a positive bias that results from the $\max$ function in Q-learning and DQN updates which outputs the maximum action value as an approximation of the maximum expected action value.\\
The Double Q-learning method was proposed in \cite{DDQN1} to alleviate this problem and later extended to DQN \cite{DDQN2} to produce the Double DQN (DDQN) method. Since Q-learning uses the same estimator to select and evaluate an action, which results in overoptimistic action values, we can interpret it as a single estimator. In Double Q-learning, the task of evaluation and selection is decoupled by using double estimator approach consisting of two functions: $Q^A$ and $Q^B$. The $Q^A$ function is updated with a value from the $Q^B$ function for the next state and the $Q^B$ function is updated with a value from the $Q^A$ function for the next state.\\
Let,
\begin{equation}\label{13}
  a^\ast=\argmax_{a_t}Q_t^A(s_{t+1},a_t)
\end{equation}
\begin{equation}\label{13}
  b^\ast=\argmax_{a_t}Q_t^B(s_{t+1},a_t)
\end{equation}
Then,
\begin{equation}\label{15}
  Q_{t+1}^A(s_t,a_t) = Q_t^A(s_t,a_t)+\eta[r_t+\gamma Q_t^B(s_{t+1},a^\ast)-Q_t^A(s_t,a_t)]
\end{equation}
\begin{equation}\label{16}
  Q_{t+1}^B(s_t,a_t) = Q_t^B(s_t,a_t)+\eta[r_t+\gamma Q_t^A(s_{t+1},b^\ast)-Q_t^B(s_t,a_t)]
\end{equation}
where, $a^\ast$ is the action with the maximum q-value in state $s_{t+1}$ according to the $Q^A$ function and $b^\ast$ is the action with the maximum q-value in state $s_{t+1}$ according to the $Q^B$ function.\\
The double estimator technique is unbiased which results in no overestimation of action values, since action evaluation and action selection is decoupled into two functions that use separate $\max$ function estimates of action values. In fact, thorough analysis of Double Q-learning in \cite{DDQN1} shows that it sometimes might underestimate action values.\\
The Double Q-learning algorithm was adapted for large state-action spaces in \cite{DDQN2} by forming the Double DQN method in a similar way as DQN. The two Q-functions ($Q^A$ and $Q^B$) can be parameterised by two sets of weights $\theta$ and $\theta^{\prime}$. At each step, one set of weights $\theta$ is used to update the greedy policy and the other $\theta^{\prime}$ to calculate it's value. For Double DQN, equation 10 can be written as:
\begin{equation}\label{17}
  y = r_t+\gamma Q(s_{t+1},\argmax_aQ(s_{t+1},a_t;\theta_i);\theta_i^{\prime})
\end{equation}
The first set of weights $\theta$ are used to determine the greedy policy just like in DQN. But, in Double DQN, the second set of weights $\theta^{\prime}$ is used for an unbiased value estimation of the policy. Both sets of weights can be updated symmetrically by switching between $\theta$ and $\theta^{\prime}$.\\
The target value network in DQN can be used as the second Q-function instead of introducing an additional network. So, the weights at the $i^{th}$ iteration are used to evaluate the greedy policy and the weights at the previous iteration to estimate it's value. The update  rule remains the same as DQN, while changing the target as:
\begin{equation}\label{18}
  y = r_t+\gamma Q(s_{t+1},\argmax_aQ(s_{t+1},a_t;\theta_i);\theta_{i-1})
\end{equation}
Note that in both DQN and DDQN, the target network uses the parameters of the previous iteration $i-1$. However, to generalise, the target network can use parameters from the any previous $(i-k)^{th}$ iteration. Then, the target network parameters are updated periodically with the copies of the parameters of the online network.

\subsection{\textbf{Dueling DQN}}
In quite a few RL applications, it is sometimes unnecessary to estimate the value of each action. In many states, the choice of action has no consequence on  the outcome. A new architecture for model-free Reinforcement Learning, called the dueling architecture, is proposed in \cite{DUELING}. The dueling architecture explicitly separates state values and action advantage values into two streams which share a common feature extraction backbone neural network. The architecture is similar to that of the DQN and DDQN architectures; the difference being that instead of a single stream of fully connected layers, there are two streams providing estimates of the value and state-dependent advantage functions. The two streams are combined at the end producing a single Q-function.\\
One stream outputs a scalar state value, while the other outputs an advantage vector having dimensionality equal to number of actions. Both the streams are combined at the end to produce the Q-function estimate. The combining module at the end can simply aggregate the value and advantage estimates as:
\begin{equation}\label{19}
  Q(s_t,a_t;\theta,\alpha,\beta)=V(s_t;\theta,\beta) + \mathcal{A}(s_t,a_t;\theta,\alpha)
\end{equation}
where, $\theta$ are the parameters of the lower layers of the neural network (before streams are split); $\alpha$ and $\beta$ are the parameters of the advantage and value function streams. However, such an aggregation of streams would require $V(s_t;\theta,\beta)$ to be replicated as many times as the dimensionality of $\mathcal{A}(s_t,a_t;\theta,\alpha)$. Also, value and advantage estimates cannot be uniquely recovered given the estimated Q-function.\\
One way of addressing these issues, proposed in \cite{DUELING}, is to force the advantage function estimator to have zero value at the selected action. This aggregation is implemented in the combining module as:
\begin{multline}\label{20}
  Q(s_t,a_t;\theta,\alpha,\beta)=V(s_t;\theta,\beta) + \\ (\mathcal{A}(s_t,a_t;\theta,\alpha)-\max_{a_{t+1}\in A}\mathcal{A}(s_t,a_{t+1};\theta,\alpha))
\end{multline}
Now, for a chosen action (action with max Q-function), $a^{\ast}=\argmax_{a_{t+1}\in A}Q(s_t,a_{t+1};\theta,\alpha,\beta)$, putting in equation 20, we get $Q(s_t,a^{\ast};\theta,\alpha,\beta)=V(s_t;\theta,\beta)$. Hence, the two streams can be uniquely recovered.\\
In \cite{DUELING}, another way of aggregation is proposed which eliminates the $\max$ operator.
\begin{multline}\label{21}
 Q(s_t,a_t;\theta,\alpha,\beta)=V(s_t;\theta,\beta) + \\ (\mathcal{A}(s_t,a_t;\theta,\alpha)-
 \frac{1}{|A|}\sum_{a_{t+1}}\mathcal{A}(s_t,a_{t+1};\theta,\alpha))
\end{multline}
where, $|A|$ is the number of actions. Even though value and advantage estimates are now off-target by a constant, this way of aggregation improves stability by capping the changes in the advantage estimates by their mean and enhances overall performance.\\
In this paper, we use above mentioned off-policy, model-free algorithms on our novel fire evacuation environment and significantly improve performance for each of the above methods by transferring tabular Q-learning knowledge of the building structure into these methods.

\section{Fire Emergency Evacuation System}
In this paper, we propose the first fire evacuation environment to train reinforcement learning agents and a new transfer learning based tabular Q-learning+DQN method that outperforms state-of-the-art RL agents on the proposed environment. The fire evacuation environment consists of realistic dynamics that simulate real-world fire scenarios. For such a complex environment, an out-of-the-box RL agent doesn't suffice. We incorporate crucial information in the agent before training it, like the shortest path to the exit from each room. The rest of the section explains the entire system in detail.

\subsection{\textbf{The Fire Evacuation Environment}}
We propose the first benchmark environment for fire evacuation to train reinforcement learning agents. To the best of our knowledge, this is the first environment of it's kind. The environment has been specifically designed to simulate realistic fire dynamics and scenarios that frequently arise in real world fire emergencies. We have implemented the environment in the OpenAI gym format \cite{GYM}, to facilitate further research.\\
The environment has a graph based structure to represent a building model. Let $G=(V,E)$ be an undirected graph, such that $V=\{v_1,v_2,...,v_n\}$ is a set of vertices that represents $n$ rooms and hallways and $E=\{e_1,e_2,...,e_m\}$ is a set of edges that represents $m$ paths connecting the rooms and hallways. A simple fire evacuation environment consisting of 5 rooms and paths connecting these rooms is shown in Fig. 2.\\
\begin{figure}
  \includegraphics[width=\linewidth]{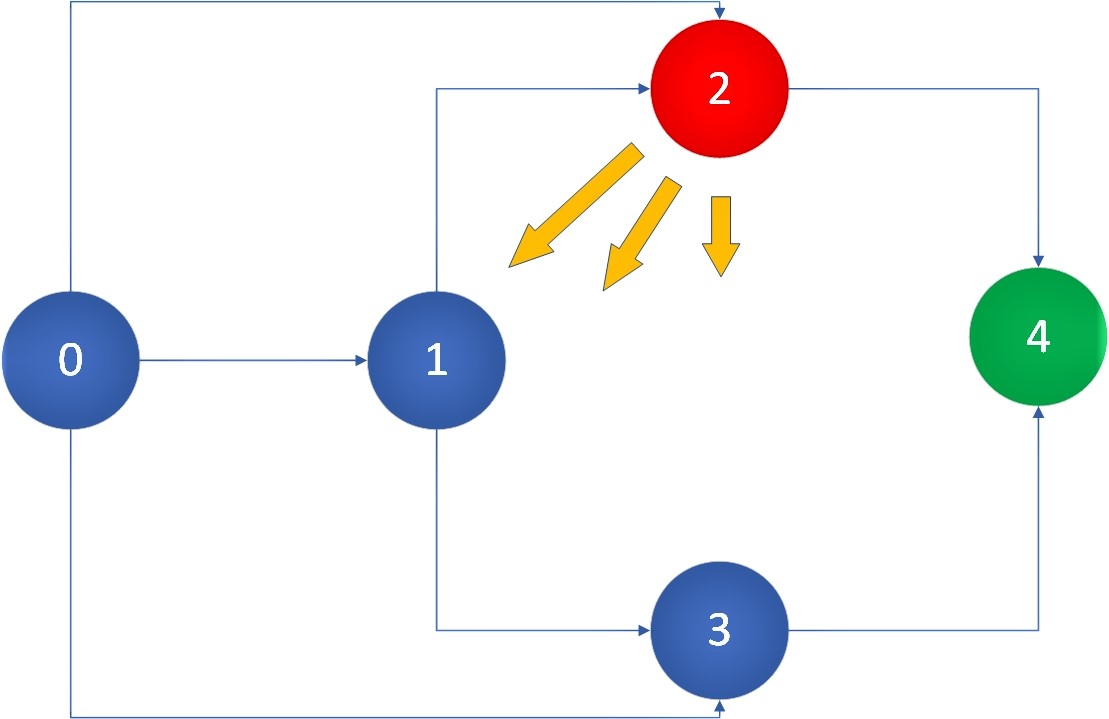}
  \label{fig:sfig1}
  \caption{A Simple Fire Evacuation Environment}
  \small
  The red vertex indicates fire in that room and the green vertex is exit. The orange arrows show the fire spread direction (more towards 1 compared to 3).
\end{figure}
To represent the graph consisting of rooms, hallways and connecting paths, we use the adjacency matrix $M_A$. It is a square matrix consisting of elements $[0,1]$ that indicate whether a pair of vertices is connected by an edge or not. The adjacency matrix is used to represent the structure of the graph and check the validity of actions performed by the agent. The adjacency matrix for the building model in Fig. 2 is given by:

\[
M_A=
  \begin{bmatrix}
    0 & 1 & 1 & 1 & 0 \\
    1 & 0 & 1 & 1 & 0 \\
    1 & 1 & 0 & 0 & 1 \\
    1 & 1 & 0 & 0 & 1 \\
    0 & 0 & 0 & 0 & 0
  \end{bmatrix}
\]

The environment dynamics are defined as follows:
\paragraph{\textbf{State}}Each vertex $v_i$ of the graph represents a room and each room is associated with an integer $N_i$, which is the number of people in that room. The state of the environment is given by a vector consisting of the number of people in each room $S=[N_1,N_2,...,N_n]$. To force the RL agent to learn the environment dynamics by itself, the environment doesn't provide any other feedback to the agent apart from the state (number of people left in each room) and the reward.
\paragraph{\textbf{Action}}An agent performs an action by moving a person from one room to the other and the state is updated after every valid action. Therefore, the action space is discrete. To keep things simple, we restrict the agent to move one person from one room at a time step. The agent can move a person from any room to any other room at any time step, even if the rooms aren't connected to each other by a path. So, the number of possible actions at each step is $n^2$.\\
This action space is necessary so that the agent can easily generalize to any graph structure. Also, this enables the agent to directly select which room to take people from and which room to send people to, instead of going through each room in a serial manner or assigning priorities.\\ 
When the agent selects an action, where there is no path between the rooms, the agent is heavily penalized. Due to this unrestricted action space and penalization, the agent is able to learn the graph structure (building model) with sufficient training and only performs valid actions at the end. The adjacency matrix is used to check the validity of actions.\\
Note that our graph based fire evacuation environment has $n^2$ possible actions (even though many of them are illegal moves and incur huge penalties), where $n$ is the number of rooms. Even for a small toy example of $n=5$ rooms, the total number of possible actions is $25$, which is a lot more than almost all of the OpenAI gym environments and Atari game environments \cite{GYM}.
\paragraph{\textbf{Reward}}We design a reward function specifically suited for our environment. We use an exponential decay function to reward/penalize the agent depending on the action it takes and to simulate fire spread as well. The reward function looks like this:
\begin{equation}\label{22}
  r(v_j,t) = -[d(v_j,t)]^t
\end{equation}
where, $t$ is the time step, $v_j$ is the room where a person is moved to and $d(.)$ is the degree of fire spread for a room. $d(.)$ returns a positive number and if a room has a higher value of degree of fire spread, that means that fire is spreading more rapidly towards that room. We explicitly assign degrees to each room using a degree vector $D=[d(v_1,t),d(v_2,t),...,d(v_n,t)]$, where the maximum value belongs to the room where the fire is located.\\
Using such a reward function ensures the following: Firstly, the reward values drop exponentially every time step as the fire increases and spreads. Secondly, the reward of an action depends on the room where a person is moved to. The reward function will penalize an action more heavily if a person is moved to a more dangerous room (higher degree of fire spread towards that room). This is because the function yields more rapidly decaying negative rewards. Lastly, the function yields a negative reward for every action which forces the agent to seek the least number of time-steps.  The reward for reaching the exit is a constant [$r(v_j=exit,t)=+10$].
\paragraph{\textbf{Fire Location(s) and Exit(s)}}The room where the fire occurs is given the highest degree, hence the maximum penalty for entering. The direction of fire spread is randomly decided and the degrees are assigned accordingly. The degrees are updated gradually to simulate fire spread.
\begin{equation}\label{23}
  d(v_j,t+1) = d(v_j,t) + \delta_j; \quad \forall v_j \quad in \quad V
\end{equation}
where, $\delta_j$ is a small number ($0 \leq \delta \leq 1$) associated with $v_j$. $\delta$ is assigned to each room according to fire spread direction. So, $\delta$ can be used to determine fire spread direction, since higher value of $\delta$ for a room means that fire is spreading towards that room more rapidly.\\
As shown in Fig. 2, the fire spread is randomly and independently decided for all rooms $v_j$.
The exit is also treated like a room. The only difference being that the agent gets a positive reward for moving people to the exit. The number of people at the exit is reset to zero after every action. The rooms which are exits are stored in a vector $\mathcal{E}$.
\paragraph{\textbf{Bottleneck}}Probably one of the most important feature in our proposed fire evacuation environment that enhances realism is the bottlenecks in rooms. We put an upper limit on the number of people that can be in a room at a time step. This restriction ensures congestion control during evacuation, which has been a huge problem in emergency situations. The bottleneck information is not explicitly provided to the agent, instead the agent learns about this restriction during training, since a negative reward is received by the agent if the number of people in a room exceed the bottleneck value. The bottleneck $\mathcal{B}$ is set to 10 in our experiments.
\paragraph{\textbf{Uncertainty}} To take into account uncertain behaviour of the crowd and introduce stochasticity in the environment, a person moves from one room to the other with probability $1-p$. This means that an action $a_t$, selected by the agent at time-step $t$, is performed with probability $1-p$ or ignored with probability $p$. If the action is ignored, then there is no change in the state, but the reward received by the agent is as if the action was performed. This acts like a regularizing parameter and due to this, the agent is never able to converge to the actual global minimum. In our experiments, the uncertainty probability $p$ is kept at $0.1$.
\paragraph{\textbf{Terminal Condition}}The terminal/goal is reached once there are no people in any of the rooms [$\sum_{i=1}^{n}N_i=0$].\\
The pseudocode for the proposed environment is given in Algorithm 1.
\begin{algorithm}[h]
 \textbf{Environment variables: $M_A$, $\mathcal{B}$, $\mathcal{E}$, $D$, $p$}\\
 \textbf{Input: $S=[N_1,N_2,...N_n]$}\\
 $t=0$\;
  \While{not Terminal}{
  $t=t+1$\;
  $a = agent.action(S)$\;
  $v_i = a\%n$\;
  $v_j = a/n$\;
  \If{$p\geq random.uniform(0,1)$}{$r = -D[v_j]^t$\;}
  \Else{
  \If{$SUM(S)==0$}{\emph{Terminal}\;}
  \ElseIf{$M_A[v_i,v_j]==1$ and $v_j$ in $\mathcal{E}$}{
   $r = +10$\;
   $S[v_i]=S[v_i]-1$\;
   }
   \ElseIf{$M_A[v_i,v_j]==0$}{
   $r = -2(\max(D)^t)$\;
  }
  \ElseIf{$S[v_j]\geq\mathcal{B}$}{
  $r = -0.5(\max(D)^t)$\;
 }
 \Else{ $r = -D[v_j]^t$\;
 $S[v_i]=S[v_i]-1$\;
 $S[v_j]=S[v_j]+1$\;
 }
	}
 Update $D$ according to $\delta$
 }
 \caption{Fire Evacuation Environment Pseudocode}
\end{algorithm}
From Algorithm 1, we can see that a heavier penalty is received by the agent for an illegal move compared to bottleneck restriction violation and moving towards fire. In a way, rewards are used to assign priorities to scenarios. It can easily be changes if needed.
\subsubsection*{\textbf{Pretraining Environment}}
We create two instances of our environment: one for fire evacuation and the other for shortest path pretraining. For the pretraining instance, we consider only the graph structure and the aim is to get to the exit from every room in the minimum number of time-steps.\\
The pretraining environment consists of the graph structure only, i.e. the adjacency matrix $M_A$. The pretraining environment doesn't contain fire, the number of people in each room or bottlenecks. The rewards are static integers: -1 for every path to force the agent to take minimum time-steps, -10 for illegal actions (where there is no path) and +1 for reaching the exit. The agent is thus trained to incorporate shortest path information of the building model.\\
The pseudocode for the pretraining environment is given in Algorithm 2.
\begin{algorithm}[h]
 \textbf{Environment variables: $M_A$, $\mathcal{E}$}\\
 \textbf{Input: $V=[v_1,v_2,...v_n]$}\\
  $t=0$\;
  $s=RandomSelection(V)$\;
  \While{not Exit}{
  $t=t+1$\;
  $a = agent.action(s)$\;
  $v_i = a\%n$\;
  $v_j = a/n$\;
  \If{$M_A[v_i,v_j]==1$ and $v_j$ in $\mathcal{E}$}{
   $r = +1$\;
   \emph{Exit}\;
   }
  \ElseIf{$M_A[v_i,v_j]==0$}{
   $r = -10$\;
  }
  \Else{ $r = -1$\;
  $s=v_j$\;
  }
 }
 \caption{Shortest Path Pretraining Environment Pseudocode}
\end{algorithm}
The procedure is repeated until the agent converges to the shortest path from any room to the exit.
\subsection{\textbf{Similarities and Differences with Other Environments}}
The fire evacuation environment is implemented in the OpenAI gym format \cite{GYM}, to enable future research on the topic. OpenAI gym environments consists of four basic methods: \emph{init}, \emph{step}, \emph{render} and \emph{reset}. Our environment consists of the same four methods.\\
The \emph{init} method consists of the initialization conditions of the environment. In our case, it contains the action space size, $A$, the state space size, $|S|$, the starting state $S$ which is an array consisting of the number of people in each room (vertex), the adjacency matrix of the graph based building model, $M_A$ and the fire location(s), $F$. The \emph{reset} method simply sets the environment back to the initial conditions.\\
The \emph{step} method is like the Algorithm 1, without the while loop. The \emph{step} method takes in the action $a_t$ performed at time-step $t$ as the argument and returns the next state $s_{t+1}$, boolean variable for terminal $T$ (indicating whether the terminal state was reached with the action performed or not) and the reward $r_{t+1}$ for performing the action. The next state, reward and terminal depend on the conditions of the environment as shown in Algorithm 1. The \emph{render} method simply returns the current state $s_t$.\\
The pretraining environment instance has the same structure of methods. The only difference is in the \emph{step} method, shown in Algorithm 2 excluding the while loop, where the reward system is changed with fewer conditions and the state $S$ is represented as the set of empty vertices (rooms with no people) of the graph.\\
Even though our environment might have the same structure as any OpenAI gym environment, it differs a lot in functionality from other environments or any game-based environments. In some ways, it might look like the mouse maze game in which the player (mouse) needs to reach the goal (cheeze) in the least possible steps through a maze. But, it is drastically different in many ways:
\begin{itemize}
	\item[$\bullet$] Our environment is a graph based environment with much less connectivity then the maze environment, which makes finding the optimal route difficult.
	\item[$\bullet$] The optimal path(s) might change dynamically from one episode to the next or within a few time-steps due to fire spread and uncertainty in the fire evacuation environment, while the optimal path(s) for the mouse maze game remains the same.
	\item[$\bullet$] All the people in all the rooms must be evacuated to the nearest exit in the minimum number of time-steps under dynamic and uncertain conditions with bottlenecks, whereas in the mouse maze environment an optimal path only from the starting point to the goal needs to be found.
	\item[$\bullet$] The fire evacuation problem is a problem in which multiple optimal paths for all people in all rooms must be found while avoiding penalizing conditions like fire, bottlenecks and fire spread, whereas the mouse maze problem is a simple point-to-point problem.
	\item[$\bullet$] The mouse maze environment is static and lacks any variations, uncertainty or dynamics. On the other hand, the fire evacuation environment is dynamic, variable and uncertain.
	\item[$\bullet$] In the maze environment, the shortest path to the goal state is always the best. But, in the fire evacuation environment, even though the DQN agent is pretrained on the shortest path information, the shortest path to the exit might not be the best due to fire, fire spread and bottlenecks.
	\item[$\bullet$] The fire evacuation environment has a much larger action space $n^2$ than the maze environment (four actions: up, down, left, right) because all actions can be performed even if they are illegal (which will yield high penalties) to make the RL agent learn the building structure (graph model).
	\item[$\bullet$] Finally, a graph is a much better way to model a building's structure than a maze, since connectivity can be better described with a graph rather than a maze. It's what graphs were made for, to depict the relationships (connections) between entities (vertices).
\end{itemize}
Hence, the fire evacuation problem is a much more complex and dramatically different problem than the mouse maze problem or any other game based problem. Even the Go game has $19\times19+1$, i.e, $362$ possible actions, whereas the fire evacuation environment has $n^2$ possible actions, i.e., as the number of rooms increase, the possible actions increase exponentially (although the Go game rules are quite complex to interpret by an RL agent).

\subsection{\textbf{Q-matrix Pretrained Deep Q-Networks}}
For the proposed graph based fire evacuation environment, we also present a new reinforcement learning technique based on the combination of Q-learning and DQN (and its variants). We apply tabular Q-learning to the simpler pretraining environment, with a small state space, to learn the shortest paths from each room to the nearest exit. The output of this stage is an $n\times n$ Q-matrix which contains q-values for state-action pairs according to the shortest path.\\
This Q-matrix is used to transfer the shortest path information to the DQN agent(s). This is done by pretraining the agent's neural network by deliberately overfitting it to the Q-matrix. After pretraining, the neural network weights have the shortest path information incorporated in them. Now, the agent is trained on the complete fire evacuation environment to learn to produce the optimal evacuation plan.\\
The main purpose of using such a strategy of training an agent by pretraining it first is to provide the agent with vital information about the environment beforehand, so that it doesn't have to learn all the complexities of the environment altogether. Since, after pretraining, the agent knows the shortest paths to the nearest exits in the building, dealing with other aspects of the environment like fire, fire spread, number of people and bottlenecks is made easier.\\
We provide two instances of our environment: simpler shortest path pretraining instance and complex fire evacuation instance. First, the agent is pretrained on the simpler instance of the environment (for shortest path pretraining) and then trained on the more complex instance (for optimal evacuation). This approach of training the agent on a simpler version of the problem before training it on the actual complex problem is somewhat similar to curriculum learning \cite{CURRICULUM}.\\
We also add a small amount of noise or offset to the Q-matrix produced by training on the pretraining environment instance. This is done by adding or subtracting (depending on the q-value) a small $\sigma$ to each element of the Q-matrix.
	\[
	Q(s,a)= 
	\begin{cases}
	Q(s,a) + \sigma,& \text{if } Q(s,a)\leq 0\\
	Q(s,a) - \sigma,& \text{if } Q(s,a)>0
	\end{cases}
	\]
where, $\sigma$ can be thought of as a regularization parameter, which is set to $10$ in our experiments. Adding noise to the Q-matrix is necessary because we don't want the DQN agent to just memorize all the paths and get stuck at a local minimum. The actual fire evacuation instance is complex, dynamic and has uncertainty which means that an optimal path at time-step $t$ might not be the optimal path at time-step $t+k$. The hyperparameter $\sigma$ acts as a regularizer.\\
Note that we add $\sigma$ if the element of the Q-matrix is negative or zero and subtract $\sigma$ if the element is positive. This is done to offset the imbalance between good and bad actions. If we just add or subtract $\sigma$ then the relative difference between q-values would remain the same. Conditional addition or subtraction truly avoids the DQN agent from being biased to a particular set of actions leading to an exit.\\   
Even though pretraining adds some overhead to the system, there are several advantages including:
\begin{description}
  \item[$\bullet$ Better Conditioning] Pretraining provides the neural network with a better starting position of weights for training compared to random initializations.
  \item[$\bullet$ Faster Convergence] Since the neural network weights are better conditioned due to pretraining, training starts closer to the optimum and hence rate of convergence is faster.
  \item[$\bullet$ Crucial Information] Especially in the case of fire evacuation, pretraining with shortest path information provides the agent with crucial information about the environment before training begins.
  \item[$\bullet$ Increased Stability] As pretraining restricts the weights in a better basin of attraction in the parameter space, the probability of divergence is reduced which makes the model stable.
  \item[$\bullet$ Fewer number of updates]As the weights are near the optimal on the error surface, the number of updates required to reach the optimum is lower, which results in fewer memory updates and requiring less training epochs.
\end{description}
The pseudocode for the proposed Q-matrix pretrained DQN algorithm is given in Algorithm 3.
\begin{algorithm}[h]
 \textbf{Environment instances:} $Pretraining\_Env()$, $Fire\_Evacuation()$\\
 \textbf{Environment variables: $M_A$, $\mathcal{B}$, $\mathcal{E}$, $D$, $V$, $S$}\\
 $S_P=Shortest\_Path()$\;
 $F_E=Fire\_Evacuation()$\;
 \Fn{$Q_{learning}()$}{
 \For{$i\gets0$ \KwTo $T_{Q}$}{
 $s_i=S_P.state()$\;
 \While{not terminal}{
 \If{$Random(0,1)<\epsilon$}{$a_i=Random\_Action()$\;}
 \Else{$a_i=\argmax(Q_i[s_i])$\;}
 $s_{i+1},r_i,terminal=agent.act(a_i)$\;
 Update $Q_{i}(s_i,a_i)$ using eq 8\;
 $s_i=s_{i+1}$\;
 }}
 \textbf{return} $Q^\ast$
 }
 \vspace{0.1cm}
 \textbf{End Function}\\
 \Fn{$Agent_{pretrain}()$}{
 $s_0=F_E.empty\_state()$\;
 \For{$j\gets0$ \KwTo $T_{Pretrain}$}{
 $O_j=agent.predict(s_0)$\;
 $L(\theta_j)=\frac{1}{2}(O_j-Q^\ast)^2$\;
 $\theta_{j+1}=\theta_j+\eta\nabla L(\theta_j)$\;
 }
 \textbf{return} $\theta^\ast$
 }
 \vspace{0.15cm}
 \textbf{End Function}\\
 \Fn{$Main()$}{
 $DQN=load\_weights(\theta^\ast)$\;
 \For{$t\gets0$ \KwTo $T$}{
 $s_t=F_E.state()$\;
 \While{not terminal}{
 $s_{t+1},r_t,terminal=DQNAgent.act(s_t)$\;
 \emph{\textbf{Train the DQNAgent by:}}\\
 \hspace{0.1cm} Calculate $L_t(\theta_t)$ using eq. 11\;
 \hspace{0.1cm} Calculate $\nabla_{\theta_t}L_t(\theta_t)$ using eq. 12\;
 \hspace{0.1cm} Update weights: $\theta_{t+1}=\theta_t+\eta\nabla L_t(\theta_t)$\;
 $s_t=s_{t+1}$\;
 }}
 \textbf{return} $DQNAgent$
 }
 \vspace{0.125cm}
 \textbf{End Function}\\
 \vspace{0.2cm}
 \caption{Q-Matrix Pretrained DQN}
\end{algorithm}
The algorithm 3 consists of 3 functions: $Q_{learning}()$ for tabular Q-learning on the pretraining environment instance for finding optimal q-values for shortest path from each room to the nearest exit; $Agent_{pretrain}()$ for overfitting the shortest path Q-matrix to incorporate the information in the DQN Agent's network; $Main()$ for using the pretrained DQN Agent to learn the optimal evacuation plan by training it on the fire evacuation environment.\\
Modifying the final training part to include Double DQN and Dueling DQN Agents is straightforward.
\subsection{\textbf{Pretraining Convergence}}
The paper \cite{TTQL} thoroughly analyses and proves conditions where task transfer Q-learning will work. We use the proved propositions and theorems from \cite{TTQL} to show that pretraining works in our case.\\
Let the pretraining instance and the fire evacuation instance be represented by two MDPs: $M_1 = \langle S,A,R_1,P_1,\gamma_1 \rangle$ for pretraining instance and $M_2 = \langle S,A,R_2,P_2,\gamma_2 \rangle$ for fire evacuation instance. So, according to proposition 1 in \cite{TTQL}:
\begin{multline}
	\Delta(M_1,M_2) = \parallel Q_1^*-Q_2^*\parallel \triangleq \frac{\parallel R_1-R_2\parallel_\infty}{1-\gamma'} \\+ \frac{\gamma''\parallel R'\parallel_\infty}{(1-\gamma'')^2}\parallel P_1-P_2\parallel_\infty + \frac{|\gamma_1-\gamma_2|}{(1-\gamma_1)(1-\gamma_2)}\parallel R''\parallel_\infty
\end{multline}
where $\Delta(M_1,M_2)$ is the distance between MDPs and $Q_1^*$ and $Q_2^*$ are the corresponding optimal Q-functions. In our case, $\gamma_1=\gamma_2$ and $P_1=P_2=1$ since our environments are deterministic MDPs, i.e., taking an action $a$ at state $s$ will always lead to a specific next state $s'$ and no other state, with probability $p(s'|s,a)=1$. This makes the second and third term of equation 24 to zero. So, the distance between the two instances of our environment is reduced to the first term only.\\
So, according to proposition 1, if the distance between two MDPs is small, then the learned Q-function from the pretraining task is closer to the optimal of the fire evacuation task compared to random initializations and hence helps in convergence to an optimum and improves the speed of convergence.\\
Also, convergence is guaranteed according to theorem 4 in \cite{TTQL}, if the safe condition is met:
\begin{equation}
	\frac{(1-\gamma)\Delta(M_1,M_2)}{BE(Q(s,a))} \leq 1
\end{equation} 
where, $BE(Q(s,a))=|[R+\gamma Q(s',a')]-Q(s,a)|$ is the Bellman error. In our case, $\Delta(M_1, M_2)$ is small and that multiplied by $1-\gamma$, which is less than $0.1$ since $\gamma>0.9$, is less than $1$. For our case, this seems obvious since the two MDPs are instances of the same MDP. This means that convergence is guaranteed, as long as the shortest path Q-matrix obtained from the pretraining environment converges.\\
Now, to prove that our method has guaranteed convergence, we need to prove that the Q-matrix is able to capture the shortest path information accurately. 
\subsection{\textbf{Convergence Analysis of Q-learning for finding shortest path}}
The guarantee of convergence for Q-learning has been discussed and proved in many different ways and for general as well as unique settings \cite{Q,CONVERGEQ}. The convergence of Q-learning is guaranteed, while using the update rule given in equation 8, if the learning rate $\eta$ is bounded between $0\leq \eta<1$ and the following conditions hold:
\begin{equation}
	\sum_{t=1}^{\infty} \eta_t=\infty, \quad \sum_{t=1}^{\infty} [\eta_t]^2 < \infty
\end{equation}
Then, $Q_t(s,a)\longrightarrow Q^*(s,a)$ as $t\longrightarrow\infty$, $\forall s,a$, with probability 1. This means that for the learning rate conditions to hold with the constraint $0\leq \eta<1$, all state-action pairs must be visited an infinite number of times. Here, the only complication is that some state-action pairs might never be visited.\\ 
In our pretraining environment, which is an episodic task, we can make sure that all state-action pairs are visited by starting episodes at random start states which is shown in Algorithm 2. Apart from this we use an $\epsilon$-greedy exploration policy to explore all state-action pairs. The initial value of $\epsilon$ and the decay rate are set according to the size of the graph based environment.\\ 
We run Q-learning on the pretraining environment for $\sim1000$ episodes so ensure that the Q-matrix converges to $Q^*(s,a)$. Since we have an action space of 25 actions for 5 rooms, running for more episodes is convenient. But, for large building models (8281 actions for the large real world building scenario, in Section 5), running for many episodes could become computationally too expensive. So, we use a type of early stopping criteria, in which we stop training the Q-matrix if there is a very small change in it's elements from one episode to the next.\\
However, as we shall see in Section 5, that we do not require early stopping at all. We are able to reduce the action space drastically and hence the Q-matrix can be trained in the same way as it was trained for smaller action spaces.\\ 
In \cite{STOCONVERGE}, the proof of convergence of Q-learning is given for stochastic processes, but in our case, the environment is deterministic. Also, in \cite{ASYNCONVERGE}, a more general convergence proof for Q-learning is provided using convergence properties of stochastic approximation algorithms and their asynchronous versions. The asymptotic bounds for the error $\triangle_t(s,a)=|Q_t(s,a)-Q^*(s,a)|$ has been shown to be bound by the number of visits to state-action pairs and $t$:
\begin{equation}
	\triangle_t(s,a) \propto \frac{1}{t^{R(1-\gamma)}}
\end{equation}
where, $R = \frac{min_{s,a}P(s,a)}{max_{s,a}P(s,a)}$ and $P(s,a)$ is the sampling probability of $(s,a)$. So, it is necessary to run the Q-learning algorithm for as many episodes as possible. Hence, we device a strategy to reduce the action space for large discrete action spaces, which are as a result of real world building models, so that it becomes feasible to train the Q-matrix for a large number of episodes.

\subsection{\textbf{Discussion on alternative Transfer Learning techniques}}
There are a few ways of pretraining an agent, some of which have been discussed and evaluated in \cite{PRETRAIN}. A naive approach would be to preload the experience replay memory with demonstration data before hand. This method, however, isn't actually pretraining. The agent trains normally with the benefit of being able to learn good transitions immediately.\\
Our method of pretraining beckons the question of pretraining the agent's network directly. Pretraining a DQN network's weights on the pretraining environment would require more time compared to tabular Q-learning. The DQN would require more time to converge. Also, in the next step where the Q-matrix is used as a fixed output to train the network's weights to overfit on the q-values requires much less time. Also, for a smaller state space (like the pretraining environment) tabular Q-learning is much more efficient than DQN. The total time taken for pretraining using the proposed method is $5.15 s$ ($3.1 s$ for tabular Q-learning and $2.05 s$ for overfitting the agent's weights on the Q-matrix) compared to $9.8 s$ for pretraining DQN directly. It's because using the direct pretraining method would effectively require the DQN to be trained twice overall (once on the pretraining environment and then on the fire evacuation environment), which is inefficient and computationally expensive.\\
Also, this complexity will grow exponentially when we train it on a large real world building model, which is shown in Section 5. For an environment with $n=91$ rooms and $8281$ actions, training a DQN agent twice would be extremely inefficient and computationally infeasible, due to the size of the neural network and computations required and the expense of backpropagation. Whereas, training the Q-matrix would only require computing equation 8.\\
One of the most successful algorithms in pretraining deep reinforcement learning is the Deep Q-learning from Demonstrations (DQfD) \cite{DQfD1,DQfD2}. It pretrains the agent using a combination of Temporal Difference (TD) and supervised losses on demonstration data in the replay memory. During training, the agent trains its network using prioritized replay mechanism between demonstration data and interactions with the environment to optimize a complex combination of four loss functions (Q-loss, n-step return, large margin classification loss and L2 regularization loss).\\
The DQfD uses a complex loss function and the drawback of using demonstration data is that it isn't able to capture the complete dynamics of the environment as it covers a very small part of the state space. Also, prioritized replay adds more overhead. Our approach is far simpler and because we create a separate pretraining instance to incorporate essential information about the environment instead of the full environment dynamics, it is more efficient than demonstration data.\\

\section{Experiments and Results}
We perform unbiased experiments on the fire evacuation environment and compare our proposed approach with state-of-the-art reinforcement learning algorithms. We test different configurations of hyperparameters and show the results with best performing hyperparameters for these algorithms on our environment. The main intuition behind using Q-learning pretrained DQN model was to provide it with important information before hand, to increase stability and convergence. The results confirms our intuition empirically.
\subsubsection*{\textbf{The Agent's Network}}
Unlike the convolutional neural networks \cite{CNNOriginal} used in DQN \cite{DQN1,DQN2}, DDQN \cite{DDQN1,DDQN2} and Dueling DQN \cite{DUELING}, we implement a fully connected feedforward neural network. The network configuration is given in Table 1. The network consists of 5 layers. The ReLU function \cite{RELU} is used for all layers, except the output layer, where a linear activation is used to produce the output.
\subsubsection*{\textbf{Environment}}
The environment given in Fig. 2 is used for all unbiased comparisons. The state of the environment is given as : $S=[10,10,10,10,0]$ with bottleneck $\mathcal{B}=10$. All rooms contain 10 people (the exit is empty), which is the maximum possible number of people. We do this to test the agents under maximum stress. The fire starts in room 2 and the fire spread is more towards room 1 than room 3 (as shown in Fig. 2 with orange arrows). Room 4 is the exit. The total number of actions possible for this environment is 25. So, the agent has to pick one out of 25 actions at each step.
\subsubsection*{\textbf{Training}}
The Adam optimizer \cite{ADAM} with default parameters and a learning rate $\eta$ of $0.001$ is used for training for all the agents. Each agent is trained for 500 episodes. Training was performed on a 4GB NVIDIA GTX 1050Ti GPU. The models were developed in Python with the help of Tensorflow \cite{TENSOR} and Keras \cite{KERAS}.
\subsubsection*{\textbf{Implementation}}
Initially, the graph connections were represented as 2D arrays of the adjacency matrix $M_A$. But, when the building model's graphs get bigger, the adjacency matrices become more and more sparse, which makes the 2D array representation inefficient. So, the most efficient and easiest way to implement a graph is as a dictionary, where the keys represent rooms and their values are an array that lists all the rooms that are connected to it.
\begin{equation*}
	dict_{graph} = \{room_i : [room_j; \quad \forall j \quad in \quad M_{A_{i,j}}=1]\}
\end{equation*}
\subsubsection*{\textbf{Comparison Graphs}}
The comparison graphs shown from Fig. 3 to Fig. 10 have the total number of time-steps required for complete evacuation for an episode on the y-axis and the number of episodes on the x-axis. The comparisons shown in the graphs are different runs of our proposed agents with exactly the same environment settings used for all the other agents as well.\\
\begin{table}[h]
\renewcommand{\arraystretch}{1.5}
\caption{Network Configuration}
\label{table1}
\centering
\begin{tabular*}{\linewidth}{|p{20mm}|p{25mm}|p{30.25mm}|}
\hline
\normalsize{\textbf{Type}} & \normalsize{\textbf{Size}} & \normalsize{\textbf{Activation}}\\
\hline
\hline
\normalsize{Dense} & \normalsize{128} & \normalsize{ReLU} \\ \hline
\normalsize{Dense} & \normalsize{256} & \normalsize{ReLU} \\ \hline
\normalsize{Dense} & \normalsize{256} & \normalsize{ReLU}\\ \hline
\normalsize{Dense} & \normalsize{256} & \normalsize{ReLU}\\ \hline
\normalsize{Dense} & \normalsize{No. of actions} & \normalsize{Linear}\\ \hline
\end{tabular*}
\end{table}
We first compare the Q-matrix pretrained versions of the DQN and its variants with the original models. The graph based comparisons between models consists of number of time-steps for evacuating all people on the $y$-axis and episode number on the $x$-axis. We put an upper-limit of 1000 time-steps for an episode due to computational reasons. The training loop breaks and a new episode begins once this limit is reached.\\
\begin{figure}
  \includegraphics[width=\linewidth]{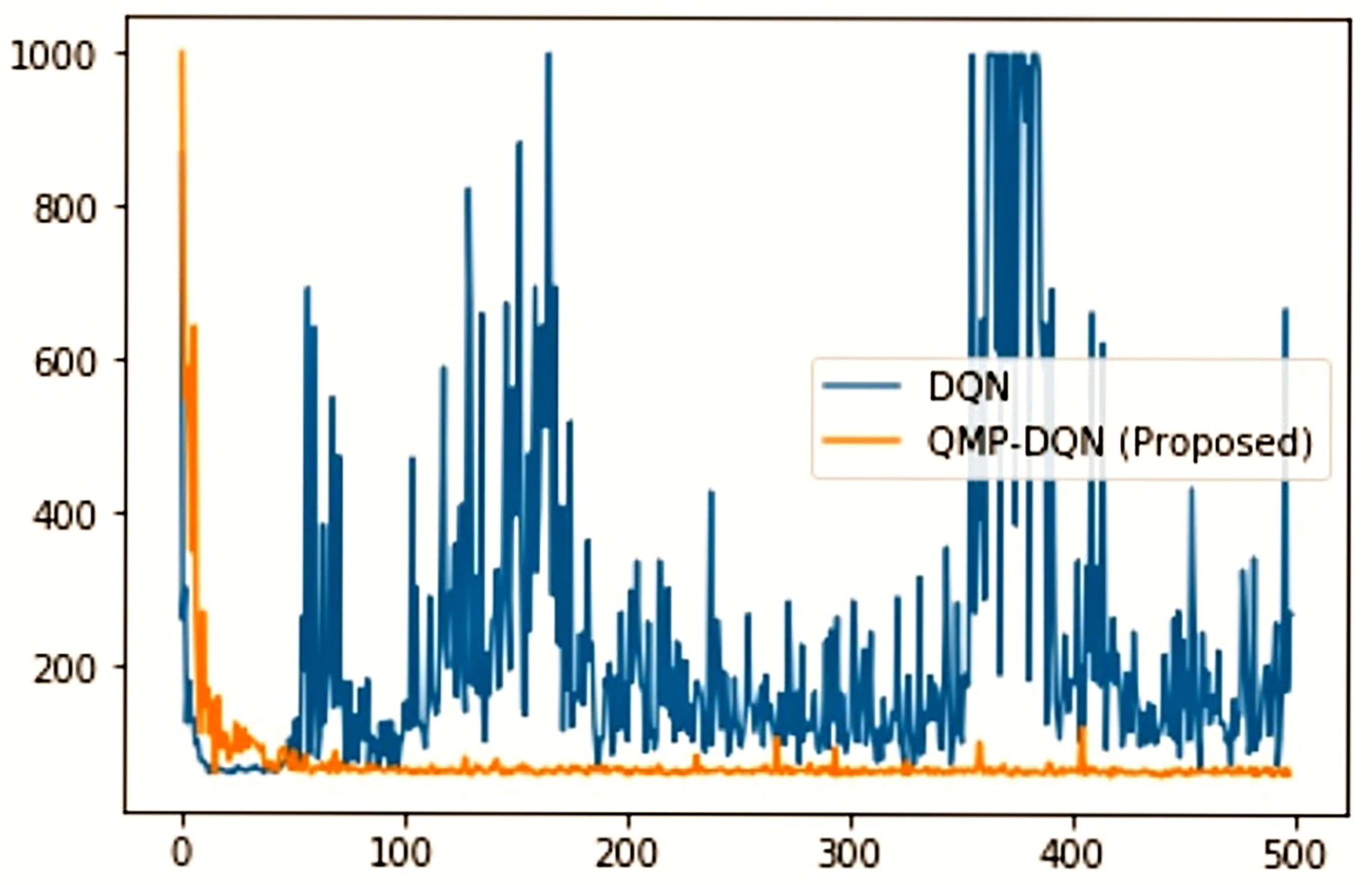}
  \label{fig:sfig1}
  \caption{Q-matrix pretrained DQN vs DQN}
\end{figure}
The graph comparing DQN with our proposed Q-matrix pretrained DQN (QMP-DQN) in Fig. 3 shows the difference in their performance on the fire evacuation environment. Although the DQN reaches the optimal number of time-steps quickly, it isn't able to stay there. The DQN drastically diverges from the solution and is highly unstable.\\
\begin{figure}
  \includegraphics[width=\linewidth]{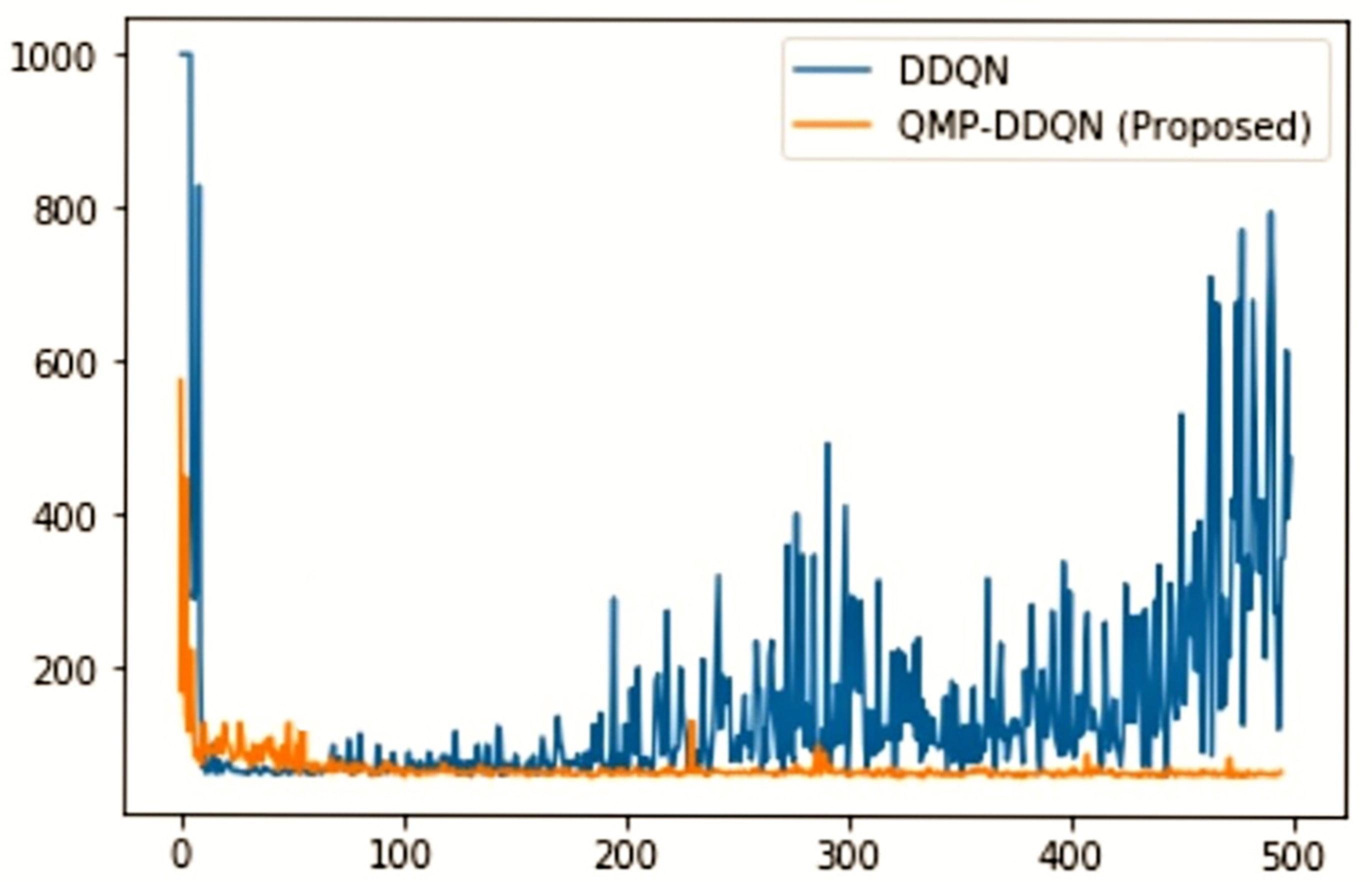}
  \label{fig:sfig1}
  \caption{Q-matrix pretrained DDQN vs DDQN}
\end{figure}
It's the same case with DDQN (Fig. 4) and Dueling DQN (Fig. 5), which, although perform better that DQN with less fluctuations and spend more time near the optimal solution. Our results clearly shows a big performance lag compared to the pretrained versions. As these results suggest that pretraining ensures convergence and stability. We show that having some important information about the environment prior to training reduces the complexity of the learning task for an agent.\\
\begin{figure}
  \includegraphics[width=\linewidth]{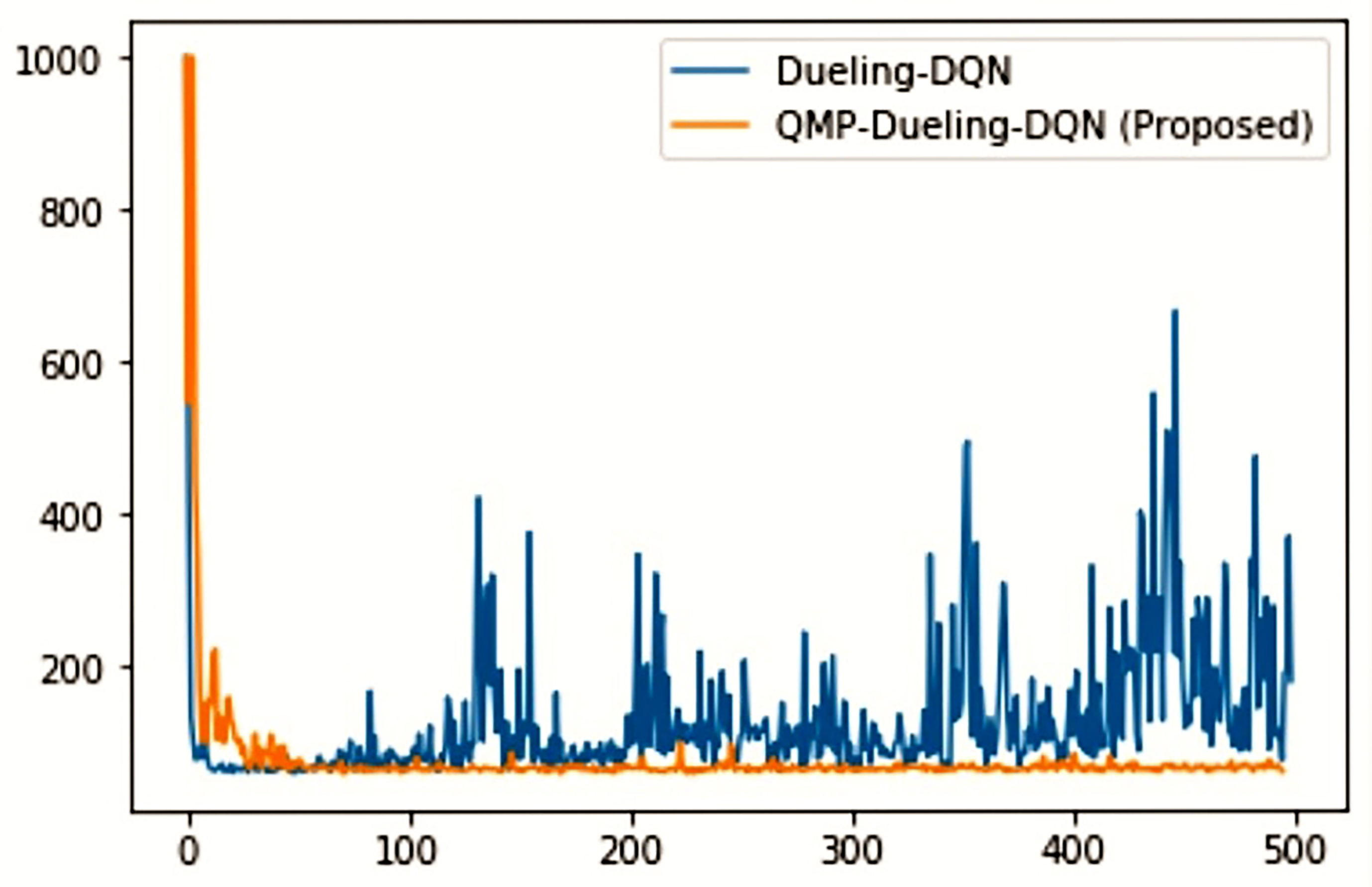}
  \label{fig:sfig1}
  \caption{Q-matrix pretrained Dueling DQN vs Dueling DQN}
\end{figure}
The original Q-learning based models aren't able to cope with the dynamic and stochastic behaviour of the environment. And since they don't posses pretrained information, their learning process is made even more difficult. Table 2 displays a few numerical results, comparing DQN, DDQN and Dueling DQN, with and without the Q-matrix pretraining on the basis of average number of time-steps for all 500 episodes, minimum number of time-steps reached during training and the training time per episode.\\
\begin{table*}[t]
\renewcommand{\arraystretch}{1.7}
\caption{Performance}
\label{table2}
\begin{tabular}{|c||c||c||c||c||c||c|}
\hline
\textbf{Model} & \textbf{Average Time-Steps} &\textbf{\makecell{Average Time-Steps \\ with Pretraining}} & \textbf{Minimum Time-Steps} & \textbf{\makecell{Minimum Time-Steps \\ with Pretraining}} & \textbf{\makecell{Training Time \\ (per episode)}} & \textbf{\makecell{Training time \\ with Pretraining \\ (per episode)}}\\
\hline
DQN & 228.2 & \textbf{76.356} & 63 & \textbf{61} & 10.117 & \textbf{6.87}\\ \hline
DDQN & 134.62 & \textbf{71.118} & 61 & \textbf{60} & 12.437 & \textbf{8.11}\\ \hline
Dueling DQN & 127.572 & \textbf{68.754} & 61 & \textbf{60} & 12.956 & \textbf{9.02}\\ \hline
\end{tabular}
\end{table*}
As it was also clear from the Figs. 3, 4 and 5, the average number of time-steps is greatly reduced with pretraining, as it makes the models more stable by reducing variance. Based on the environment given in Fig. 2, the minimum possible number of time-steps is 60. All the DQN based models are able to come close to this, but pretraining pushes these models further and achieves the minimum possible number of time-steps. Even though the difference seems small, in emergency situations even the smallest differences could mean a lot at the end. The training time is also reduced with pretraining, as the number of time-steps taken during training is reduced and pretrained models get a better starting position nearer to the optimum.\\
\begin{figure}
  \includegraphics[width=\linewidth]{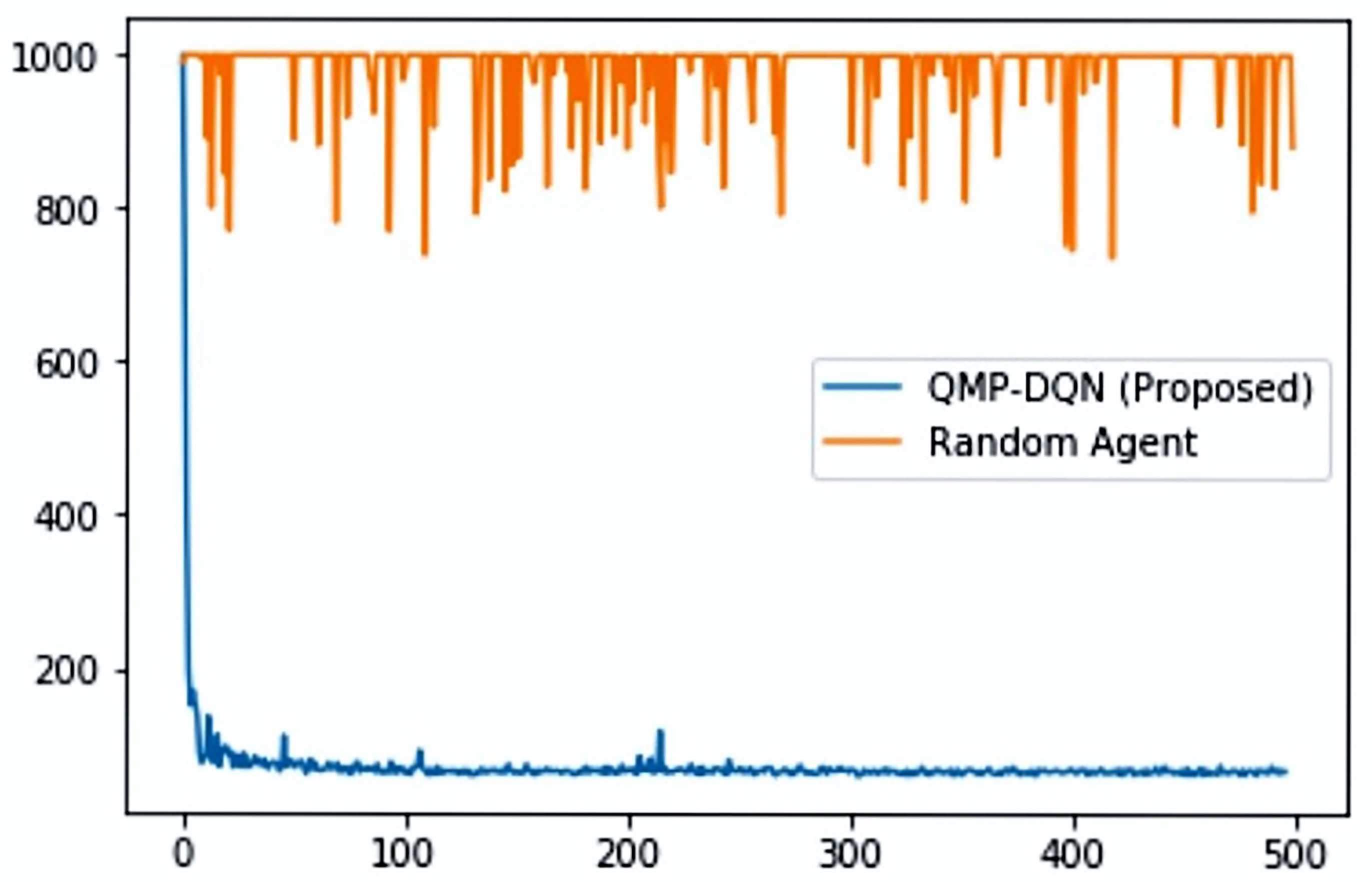}
  \label{fig:sfig1}
  \caption{Proposed method vs Random Agent}
\end{figure}
Next, we make comparisons between our proposed approach and state-of-the-art reinforcement learning algorithms. For these comparisons, we use the Q-matrix pretrained Dueling DQN model, abbreviated QMP-DQN. We also compare it with a random agent, shown in Fig. 6. The random agent performs random actions at each step, without any exploration. The random agent's poor performance of 956.33 average time-steps shows that finding the optimal or even evacuating all the people isn't a simple task.\\
\begin{figure}
  \includegraphics[width=\linewidth]{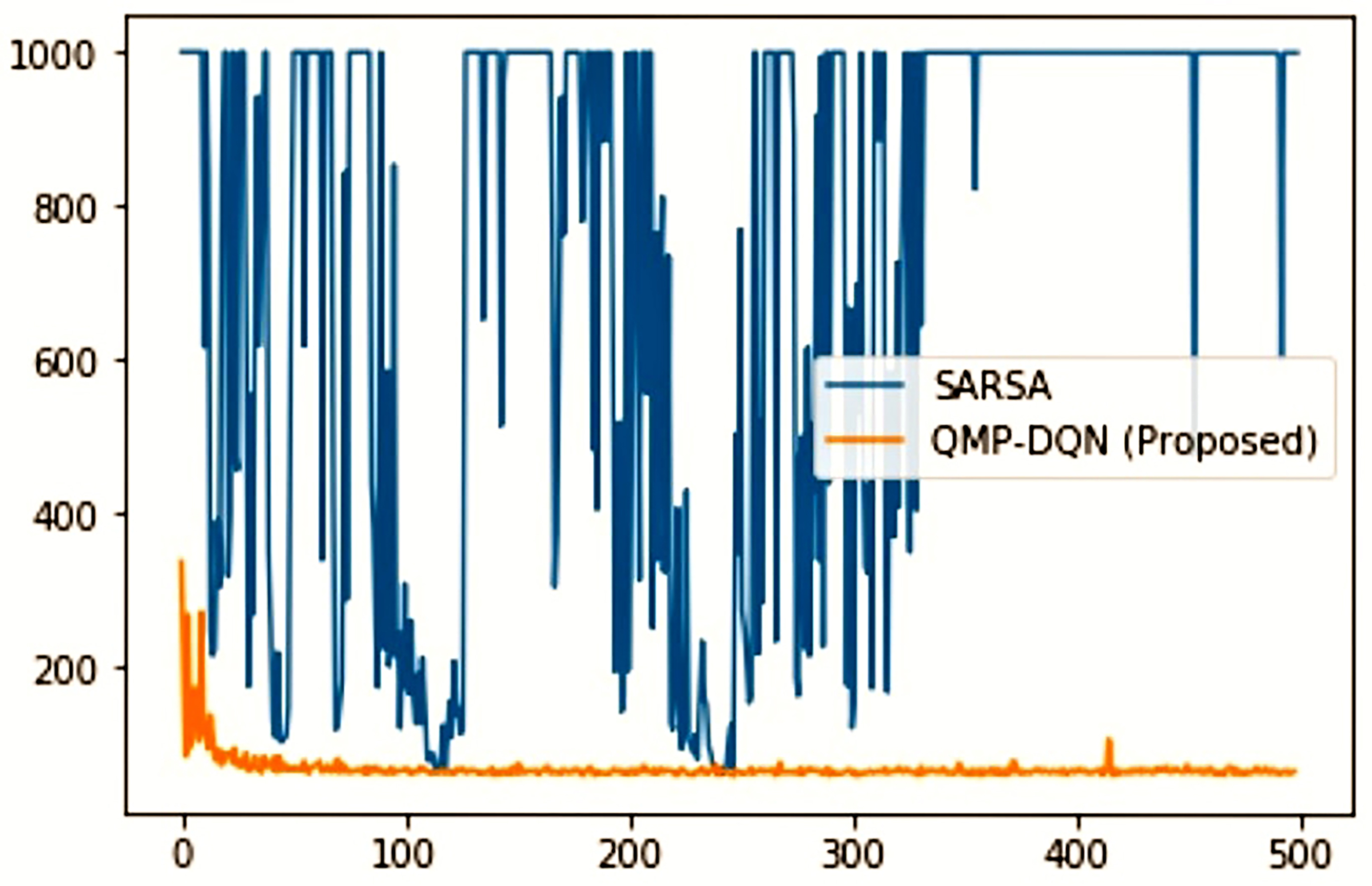}
  \label{fig:sfig1}
  \caption{Proposed method vs State-Action-Reward-State-Action method}
\end{figure}
The State-Action-Reward-State-Action (SARSA) algorithm is an on-policy reinforcement learning agent introduced in \cite{SARSA}. While Q-learning follows a greedy policy, SARSA takes the policy into account and incorporates it into its updates. It updates values by considering the policy's previous actions. On-policy methods like SARSA have a downside of getting trapped in local minima if a sub-optimal policy is judged as the best. On the other hand, off-policy methods like Q-learning are flexible and simple as they follow a greedy approach. As it is clear from Fig. 7, that SARSA behaves in a highly unstable manner and isn't able to reach the optimal solution and shows high variance.\\
\begin{figure}
  \includegraphics[width=\linewidth]{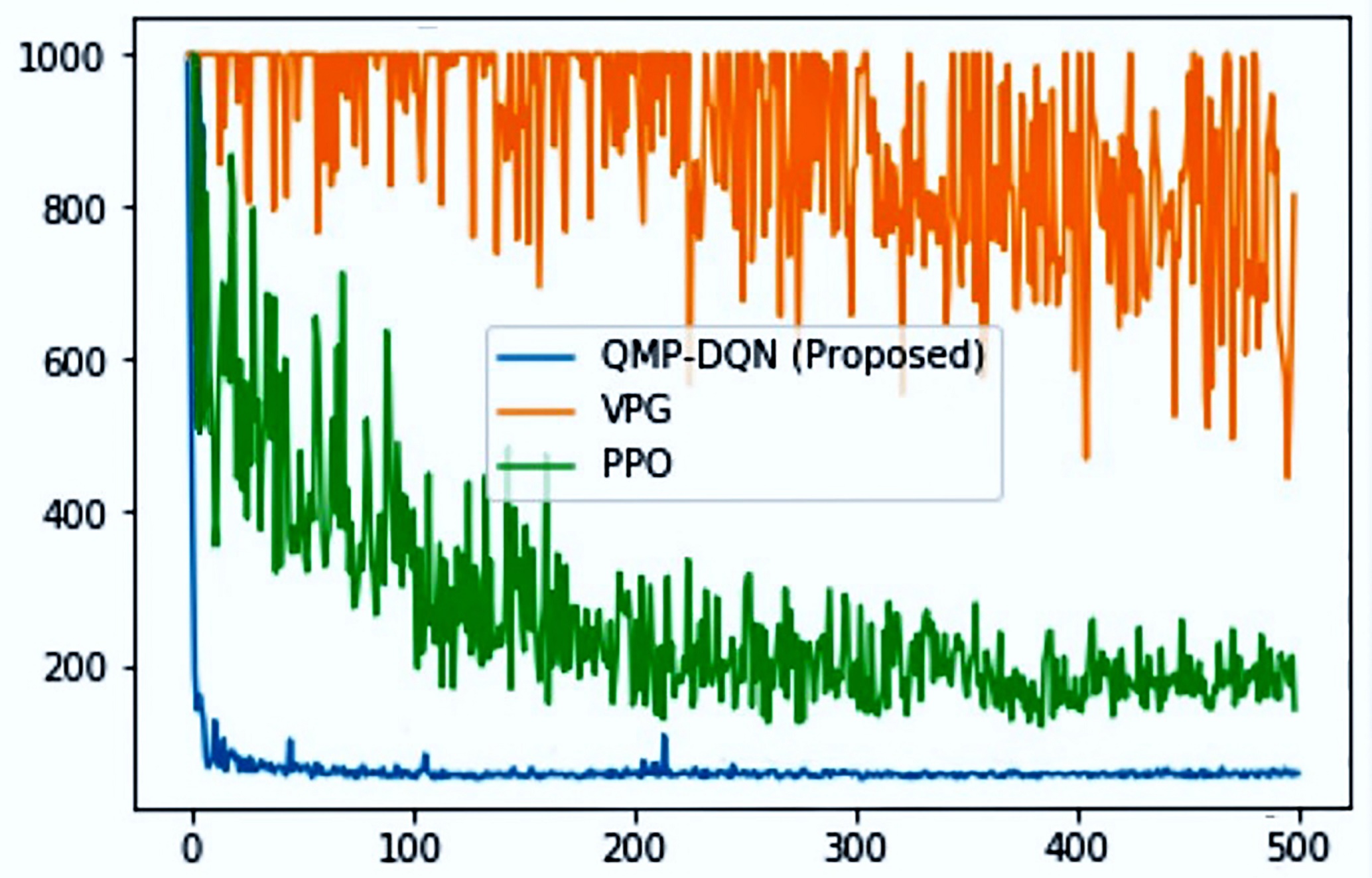}
  \label{fig:sfig1}
  \caption{Proposed method vs Policy based methods (PPO and VPG)}
\end{figure}
Policy gradient methods are highly preferred in many applications, however they aren't able to perform optimally on our fire evacuation environment. Since the optimal policy could change in a few time-steps in our dynamic environment, greedy action selection is probably the best approach. An evacuation path that seems best at a particular time step could be extremely dangerous after the next few time-steps and a strict policy of routing cannot be followed continuously due to fire spread and/or bottleneck. These facts are evident from Fig. 8, where we compare our approach to policy gradient methods like Proximal Policy Optimization (PPO) \cite{PPO} and Vanilla Policy Gradient (VPG) \cite{VPG}. Even though PPO shows promising movement, it isn't able to reach the optimum.\\
\begin{figure}
  \includegraphics[width=\linewidth]{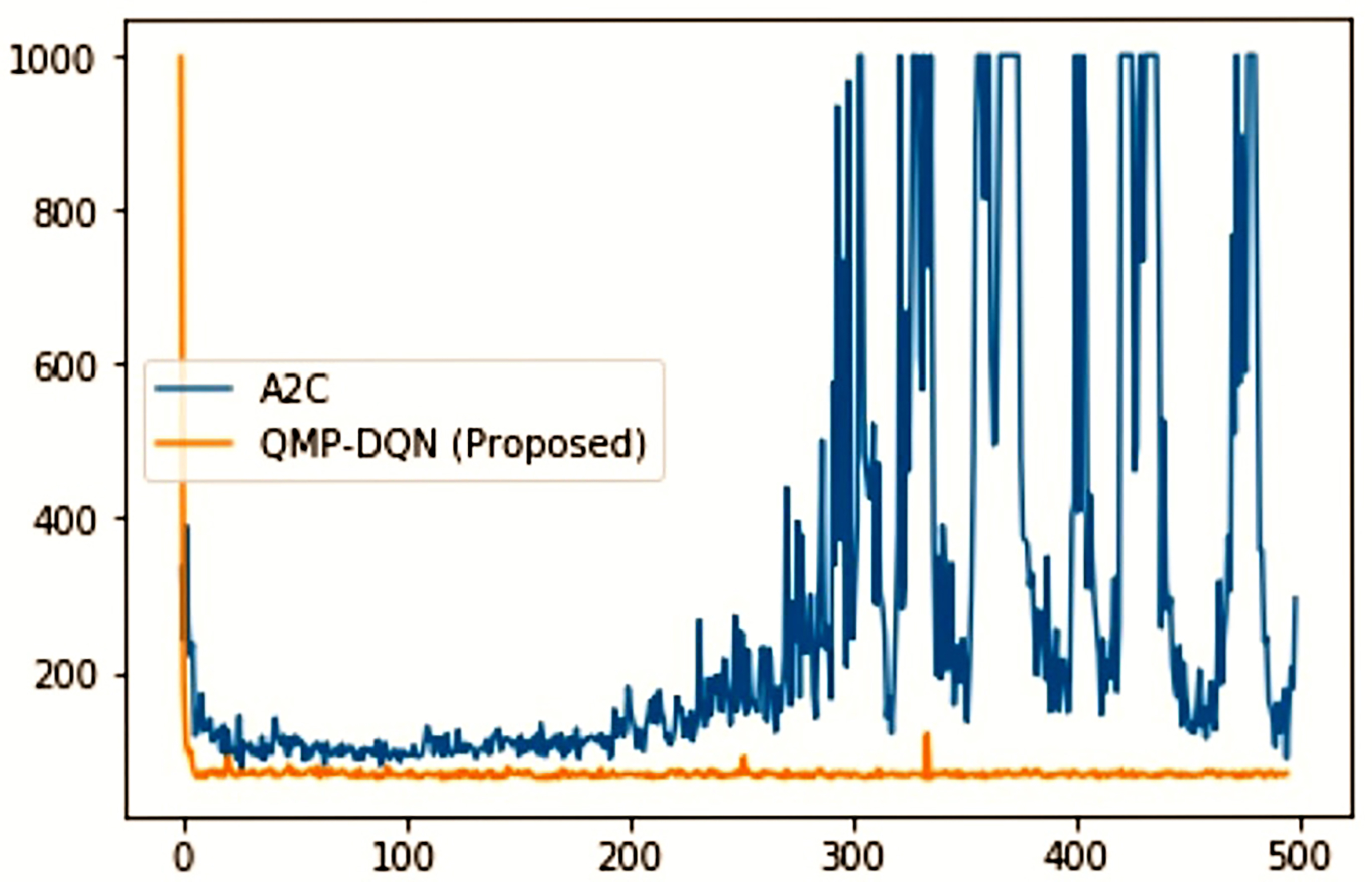}
  \label{fig:sfig1}
  \caption{Proposed method vs Synchronous Advantage Actor Critic method (A2C)}
\end{figure}
\begin{figure}
	\includegraphics[width=\linewidth]{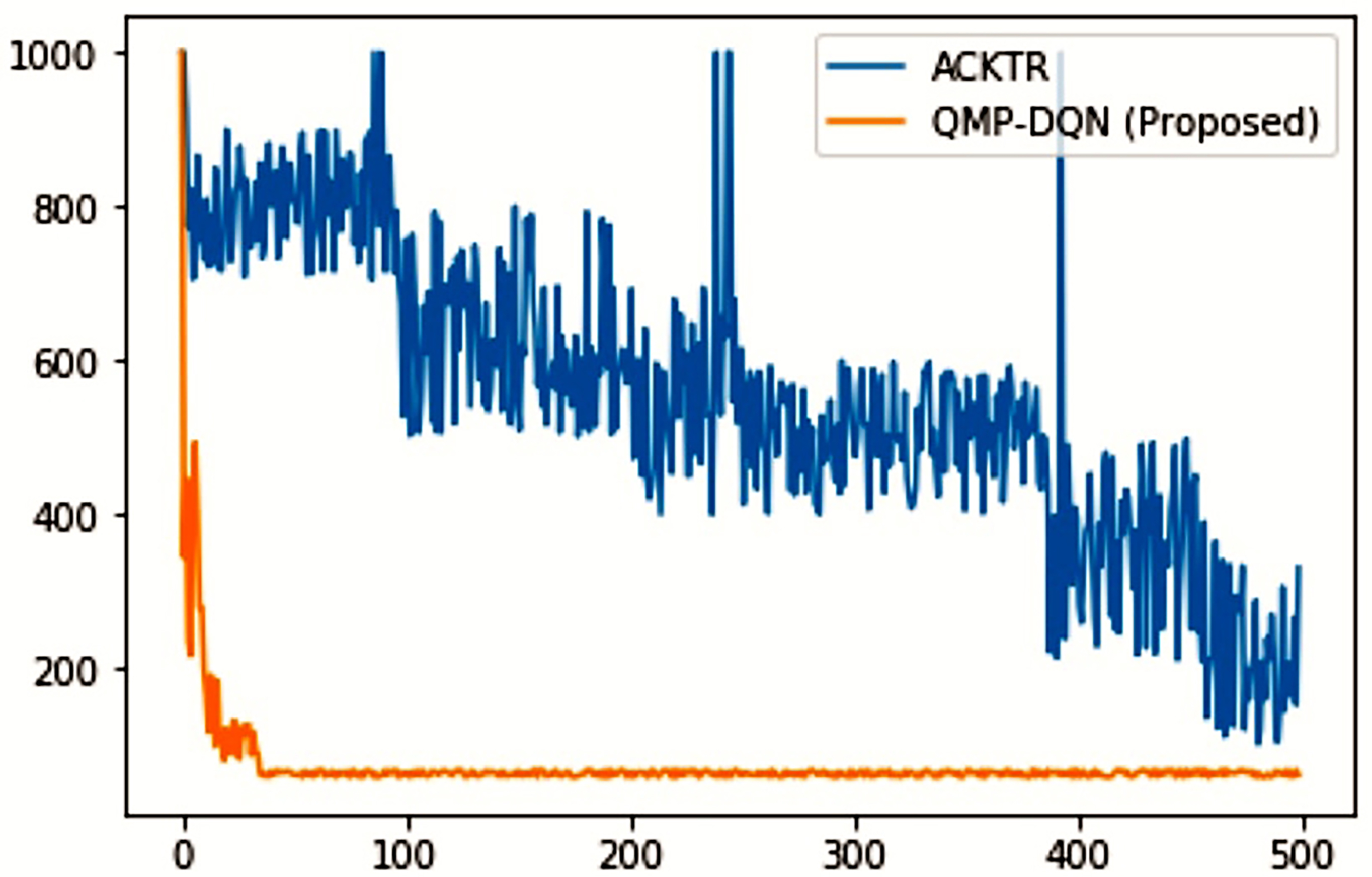}
	\label{fig:sfig1}
	\caption{Proposed method vs Actor Critic using Kronecker-Factored Trust Region (ACKTR)}
\end{figure}
\begin{table*}[h]
\renewcommand{\arraystretch}{1.8}
\caption{Comparison with State-of-the-art RL Algorithms}
\label{table3}
\centering
\begin{tabular}{|c||c||c||c|}
\hline
\textbf{Model} & \textbf{Average Time-Steps} & \textbf{Minimum Time-Steps} & \textbf{\makecell{Training Time \\ (per episode)}}\\
\hline
SARSA & 642.21 & 65 & 19.709 \\ \hline
PPO & 343.75 & 112 & 16.821 \\ \hline
VPG & 723.47 & 434 & 21.359\\ \hline
A2C & 585.92 & 64 & 25.174\\ \hline
ACKTR & 476.56 & 79 & 29.359\\ \hline
Random Agent & 956.33 & 741 & - \\ \hline
\makecell{QMP-DQN \\ (Dueling DQN Backbone)} & \textbf{68.754} & \textbf{60} & \textbf{9.02} \\ \hline
\end{tabular}
\end{table*}
Another major type of reinforcement learning algorithms are the actor-critic methods. It is a hybrid approach consisting of two neural networks: an actor which controls the policy (policy based) and a critic which estimates action values (value based). To further stabilize the model, an advantage function is introduced which gives the improvement of an action compared to an average action used in a particular state. Apart from the previously mentioned shortcomings of using policy based methods on the fire evacuation environment, the advantage function would have high variance since the best action at a particular state could change rapidly leading to unstable performance. This is apparent from Fig. 9, where we compare the synchronous advantage actor critic method (A2C) \cite{A2C} with our proposed method. The A2C gives near optimal performance in the beginning but diverges and rapidly fluctuates.\\
We do not compare our proposed method with the asynchronous advantage actor critic method (A3C) \cite{A3C}, because A3C is just an asynchronous version of A2C, which is more complex as it creates many parallel versions of the environment and gives relatively the same performance, but is not as sample efficient as claimed in \cite{NOA3C}. The only advantage of A3C is that it exploits parallel and distributed CPU and GPU architectures which boosts learning speed as it can update asynchronously. However, the main focus of this paper is not learning speed. Hence, we think that the comparison with A2C is sufficient for actor-critic models.\\
Probably the best performing Actor Critic based model is the ACKTR (Actor Critic with Kronecker-factored Trust Region) \cite{ACKTR}. The algorithm based on applying trust region optimization using Kronecker-factored approximation, which is the first scalable trust region natural gradient method for actor critic models that can be applied to both continuous and discrete action spaces. The Kronecker-factored Approximate Curvature (K-FAC) \cite{KFAC}, is used to approximate the Fisher Matrix to perform approximate natural gradient updates. We compare our method to the ACKTR algorithm, shown in Fig. 10. The results suggest that the ACKTR is not able to converge (within 500 episodes, due to slow convergence rate) and is susceptible to the dynamic changes in the environment as evident from the fluctuations. ACKTR is far too complex compared to our proposed method, which converges much faster and deals with the dynamic behaviour of the fire evacuation environment efficiently.\\
We summarize our results in Table 3. All the RL agents use the same network configuration mentioned in Table 1 for unbiased comparison. The training time for the QMP-DQN is much lower compared to other algorithms because pretraining provides it with a better starting point, so it requires less number of time-steps and memory updates to reach the terminal state. Also, SARSA and A2C come really close to the minimum number of time-steps, but as the average number of time-steps suggests, they aren't able to converge and exhibit highly unstable performance. Our proposed method, Q-matrix pretrained Dueling Deep Q-network gives the best performance on the fire evacuation environment by a huge margin.\\
Note that, in all the comparison graphs, our proposed method comes close to the global optimum, but isn't able to completely converge to it. This is because of the uncertainty probability $p$, which decides whether an action is performed or not and is set to $0.15$. This uncertainty probability is used to map the uncertain crowd behaviour. Even though, $p$, does not allow complete convergence, it also prevents the model from memorizing an optimal path which might change as the fire spreads.
\subsection*{\textbf{Multiple Fires Scenario}}
Now that we have shown that the proposed method is able to outperform state-of-the-art reinforcement learning algorithms, we test our model on a more complex and difficult environment setup. The environment configuration consists of multiple fires in different rooms and a more complex graph structure consisting of 8 rooms. The environment is shown in Fig. 10. The green node is the exit, the red nodes are the rooms where the fire is located and the orange arrows depict the direction of fire spread.\\
As we can see from Fig. 10, the fire spreads in different directions from different fire locations. This makes things especially difficult because as the fire spreads, the paths to the exit could be blocked. We do not change the configuration of our approach, except the output layer of the network, since the number of possible actions is 64 now. The State of the environment and Bottleneck given as input is: $S=[10,10,10,10,10,10,10,0]$ and $\mathcal{B}=10$.\\
\begin{figure}
  \includegraphics[width=\linewidth]{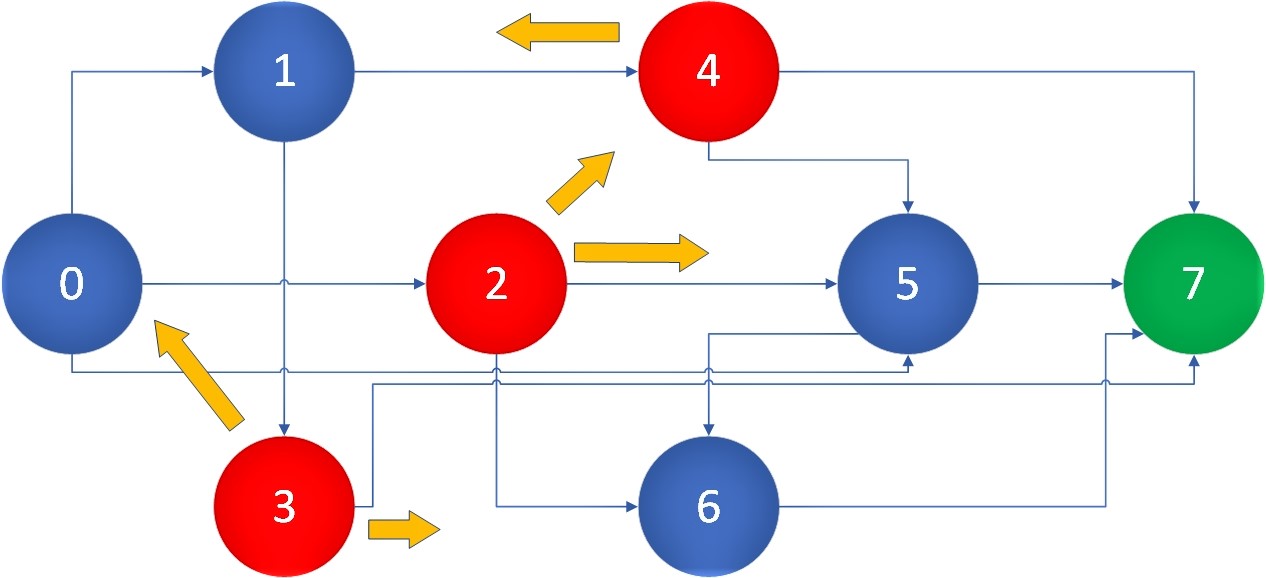}
  \label{fig:sfig1}
  \caption{A Multiple Fire Evacuation Environment}
\end{figure}
We employ the Q-matrix pretrained Dueling DQN model. Fig. 11 shows the graphical results on the multiple fires scenario. The initial fluctuations are due to $\epsilon$-greedy exploration. Since this configuration of the environment is bigger and more complex, the agent explores the environment a little longer.\\
\begin{figure}
  \includegraphics[width=\linewidth]{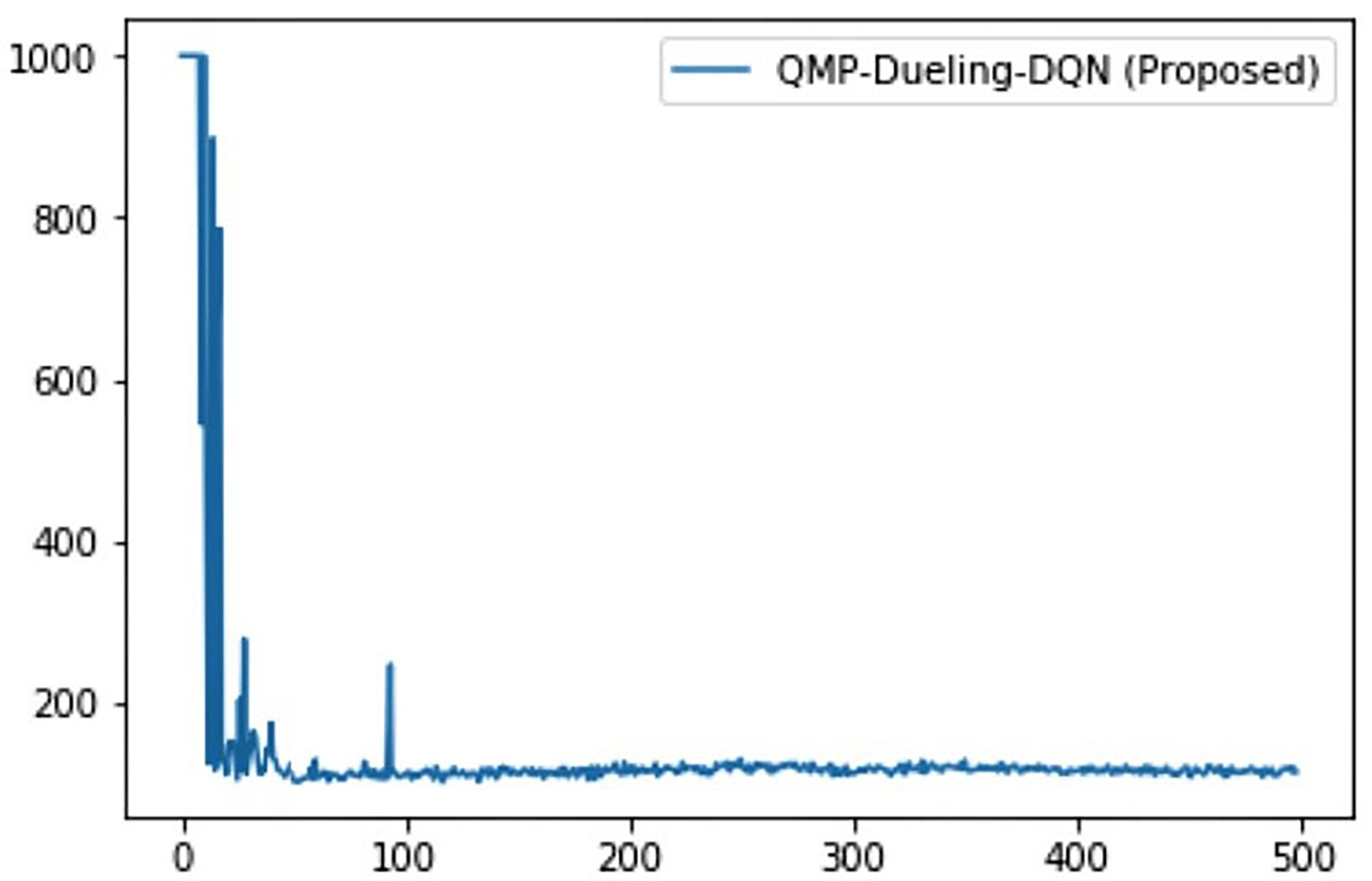}
  \label{fig:sfig1}
  \caption{Q-matrix pretrained Dueling DQN in Multiple Fire Scenario}
\end{figure}
As the results suggest from Fig. 11, the proposed model is able to converge very quickly. A few metrics for the proposed method on the multiple fires environment is given below:
\begin{description}
  \item[$\bullet$ Average number of time-steps:]119.244
  \item[$\bullet$ Minimum number of time-steps:]110
  \item[$\bullet$ Training time (per episode):]15.628
\end{description}
Note that, there is a difference of $\approx9$ time-steps between the minimum number of time-steps and average number of time-steps. This is because the average of all 500 episodes is taken which includes the initial fluctuations due to exploration and the uncertainty probability $p$.

\section{Scalability: Large and Complex Real World Scenario - University of Agder Building}
To prove that our method is capable of performing on large and complex building models, we simulate a real world building, i.e., the University of Agder A, B, C and D blocks, and perform evacuation in case of fire in any random room(s).\\
This task is especially difficult because of the resulting complex graph structure of the building and the large discrete action space. We consider the A, B, C and D blocks which are in the same building. The total number of rooms in this case is $n=91$, which means that the number of all possible actions is $8281$. This discrete action space is many times larger than any other OpenAI gym environment or Atari game environments \cite{GYM}. Even the Go game has $19\times19+1$, i.e., $362$ actions.\\
Dealing with such a large action space would require a huge agent model or moving towards to a multi-agent approach and dividing the environment into subsets, with each sub-environment for each agent to deal with. These techniques for dealing with the large discrete action space would be computationally complex and difficult to implement for the fire evacuation environment.\\
Another way could be to use a policy gradient method which are much more effective in dealing with large action spaces compared to value based methods. But, dealing with such large action spaces would require an ensemble of neural networks and tree search algorithms like in \cite{ALPHAGO} or extensive training from human interactions like in \cite{ALPHAGOZERO}. However, in a fire emergency environment we obviously can't have human interactions and we would like to solve the issue of large action space without having to use dramatically huge models. Also we saw in the previous section that even though PPO performs much better compared to other algorithms, it wasn't able to outperform our QMP-DQN methods.\\
In \cite{LDAS}, a new method to deal with extremely large discrete action spaces ($\sim$1 million actions) was proposed. The novel method, called the Wolpertinger policy algorithm, uses a type of actor-critic architecture, in which the actor proposes a proto-action in an action embedding space from which $k$ most similar actions are selected using the k-nearest neighbour algorithm. These $k$ actions are received by the critic which makes a greedy selection based on the learned q-values. This technique shows promising results, however, it is highly complex.\\
We propose a much simpler approach to deal with large number of actions. Our method consists of two stages: One-Step Simulation (OSS) of all actions resulting in an action importance vector $A_I$ and then element-wise addition with the DQN output for training. We explain our method in the following subsections.
\subsection{\textbf{One-Step Simulation and Action Importance Vector}}
We make use of the pretraining environment instance shown in Algorithm 2 to calculate the action importance vector $A_I$, as shown in Algorithm 4. The $one\_step\_sim(s,k)$ function is implemented in the environment itself to enable the environment object to use the method and the function to use the environment variables.\\
The $one\_step\_sim(s,k)$ function simulates all possible actions for each state/room for one time-step in the pretraining environment. It stores all rewards received for these actions taken from room $s$ and returns the $k$ best actions for each room $s$ which yield the $k$ highest rewards.\\
The $one\_step\_sim(s,k)$ function is run for each room $s$ in $N$, which is the total number of rooms in the environment. The equation $x[j] \longleftarrow env.one\_step\_sim(j,k)*N+j$, is used to convert the $k$ best actions returned by $one\_step\_sim(s,k)$ function for all rooms $s$, into a single vector of actions. This is necessary because the DQN agent can take any appropriate action from any room at a particular time-step. So, it outputs a single vector consisting of Q-values for all actions at each time-step.\\
\begin{algorithm}[h]
	\textbf{Environment instances:} $Pretraining\_Env()$\\
	\textbf{Environment variables:} $N\longleftarrow$ Number of rooms\\
	$env = Pretraining\_Env()$\;
	\Fn{$one\_step\_sim(s,k)$}{
	\For{$i$ \textbf{in} $action\_space$}{$s_{t+1},r,terminal = env.step(i)$\;
	$rewards[i] \longleftarrow r$\;}
	\textbf{return} $rewards.argsort(k)$\;
	}
\vspace{0.071cm}
	\textbf{End Function}\\
	\For{$j$ \textbf{in} $N$}{$x[j] \longleftarrow env.one\_step\_sim(j,k)*N+j$\;
	}
	\For{$l$ \textbf{in} $action\_space$}{
	\If{$l$ \textbf{in} $x$}{$A_I[l] = 0$\;}
	\Else{$A_I[l] = -9999$\;}
	}
	\vspace{0.15cm}
	\caption{One-Step Simulation and $A_I$}
\end{algorithm}
After we have a unique index for all selected actions in the environment, we form the action importance vector $A_I$ by placing $0$ at index $l$, if the $l^{th}$ action is present in the vector $x$, which consists of all the $k$ best actions for each room $s$, otherwise, a large negative number (like $-9999$) at index $l$.\\
The action importance vector can be though of as a fixed weight vector which contains weight $0$ for good actions and a large negative weight for others. $A_I$ is then added element-wise with the output of the DQN $\hat{Q}$ to produce the final output $Q^*$ on which the DQN is trained on.
\begin{equation}
	Q^*=\hat{Q} \oplus A_I
\end{equation}
This makes the Q-values of the good actions to remain the same and reduces the Q-values of other actions to huge negative numbers. This method effectively reduces the action space from $O(N^2)$ to $O(kN)$, where $k\ll N$. In our experiments, we set the hyperparameter $k$ as the maximum degree of vertices in the building model's graph, i.e. $k=9$. So, in our model, the action space is effectively reduced from $8281$ actions to $819$ actions, which is a $90.1\%$ decrease.\\
Hence, our complete method consists of shortest path pretraining using Q-matrix transfer learning and action space reduction by one-step simulation and action importance and finally DQN based model training and execution. The shortest path pretraining provides the model with global graph connectivity information and the one-step simulation and action importance delivers local action selection information.\\
The action importance vector can also be thought of as an attention mechanism \cite{LAS,ATTEND1,ATTEND2,ATTEND3}. Most of the attention mechanisms employ a neural network or any other technique to output an attention vector which is then combined with the input or an intermediate output to convey attention information to a model. Unlike these methods, our proposed model combines the action importance vector with the output of the DQN. This means that the current action selection is based on a combination of the Q-values produced by the DQN and the action importance vector, but the training of the DQN is impacted by the attention vector in the next iteration of training, as the final output of the $i^{th}$ iteration is used as the label for training the model at the $i+1^{th}$ iteration.\\
One major advantage of such an attention mechanism used in our method is that, since the graph based environment has a fixed structure, the attention vector needs to be calculated just once at the beginning. We test our method on the University of Agder (UiA), Campus Grimstad building with blocks A, B, C and D consisting of $91$ rooms.\\
Note that, unlike the usual attention based models, we do not perform element-wise multiplication of the attention vector with the output of a layer. Instead, we add the attention vector because initially the DQN model will explore the environment and will have negative Q-values for almost all actions (if not all). This means that if we use a vector of ones and zeros for good and bad actions respectively and multiply element-wise with the output of a layer then, the Q-values of good actions will be copied as it is and the Q-value of other actions will    
\onecolumn
\begin{figure}[htbp]
	\centering
	\includegraphics{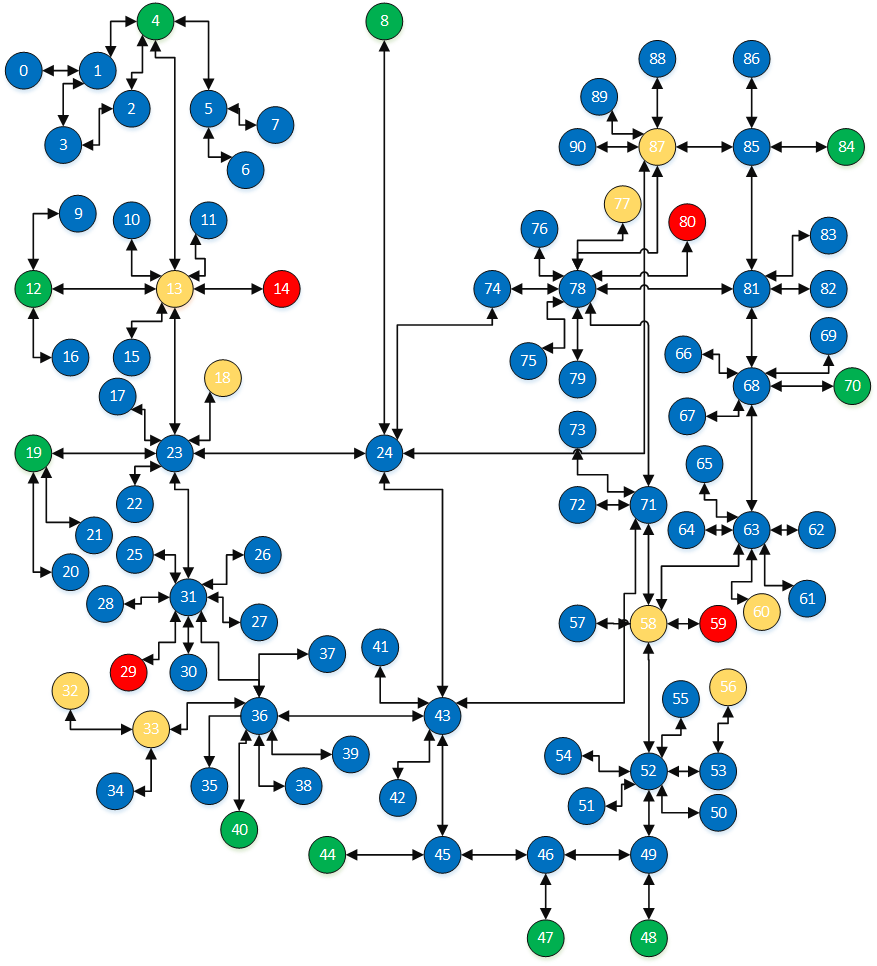}
	\caption{University of Agder Graph}
	\label{fig5}
	\small
	The red vertices indicate fire in that room and the green vertices are exits. The yellow vertices show the fire spread towards that room.
\end{figure}
\twocolumn
become zero. If the Q-value of good actions is negative in the beginning due to exploration (and lack of learning since it is the beginning of training), then the max function in the Q-value selection equation will select bad actions since they are zeros and good actions are negative. This will lead to catastrophic behaviour of the system and it will never converge. So, instead we use addition with zeros for good actions so that they remain the same and with large negative numbers for other actions so that their Q-values become so low that they are never selected.
\subsection{\textbf{Fire Evacuation in the UiA building}}
The graph for UiA's building model is based on the actual structure of the 2nd floor of blocks A, B, C and D\footnote{UiA building map can be found here: \url{https://use.mazemap.com/#v=1&zlevel=2&left=8.5746533&right=8.5803711&top=58.3348318&bottom=58.3334208&campusid=225}}. The graph for the building model is shown in Fig. 13. It consists of $91$ rooms (from room $0$ to room $90$) out of which there are $10$ exits. We simulate the fire evacuation environment in which there are multiple distributed fires in rooms $14$, $29$, $59$ and $80$. The fire spread for each fire is individually simulated in a random direction as shown by the yellow nodes in the graph.\\
As shown in Fig. 13, the  building connectivity can be quite complex and there has been limited research work that deals with this aspect. The graph structure shows that these connections between rooms cannot possibly be captured by a grid based or maze environment.\\
Also, note that, the double sided arrows in the graph enable transitions back and forth between rooms. This makes the environment more complicated for the agent since the agent could just go back and forth between 'safe' rooms and get stuck in a loop and may never converge. This point makes pretraining even more indispensable.\\
Since, the proposed method is able to reduce the action space by a lot, the neural network doesn't need to be made too large. The network configuration is given in Table 4. Note that the addition layer does not require any trainable parameters.
\begin{table}[h]
	\renewcommand{\arraystretch}{1.5}
	\caption{Network Configuration}
	\label{table1}
	\centering
\begin{tabular*}{\linewidth}{|p{25mm}|p{20mm}|p{30.25mm}|}
	\hline
	\normalsize{\textbf{Type}} & \normalsize{\textbf{Size}} & \normalsize{\textbf{Activation}}\\
	\hline
	\hline
	\normalsize{Dense} & \normalsize{512} & \normalsize{ReLU} \\ \hline
	\normalsize{Dense} & \normalsize{1024} & \normalsize{ReLU} \\ \hline
	\normalsize{Dense} & \normalsize{1024} & \normalsize{ReLU}\\ \hline
	\normalsize{Dense} & \normalsize{1024} & \normalsize{ReLU}\\ \hline
	\normalsize{Dense} & \normalsize{8281} & \normalsize{Linear}\\ \hline
	\normalsize{Addition} & \normalsize{-} & \normalsize{-}\\ \hline
\end{tabular*}
\end{table}
The neural network is trained using the Adam optimizer \cite{ADAM} with default hyperparameter settings and a learning rate $\eta=0.001$ for $5000$ episodes. The training was performed on the NVIDIA DGX-2. The optimal number of steps for evacuation in the UiA building graph is around $\sim2000$.
\subsection{\textbf{Results}}
\begin{figure*}[t]
	\includegraphics[width=\textwidth]{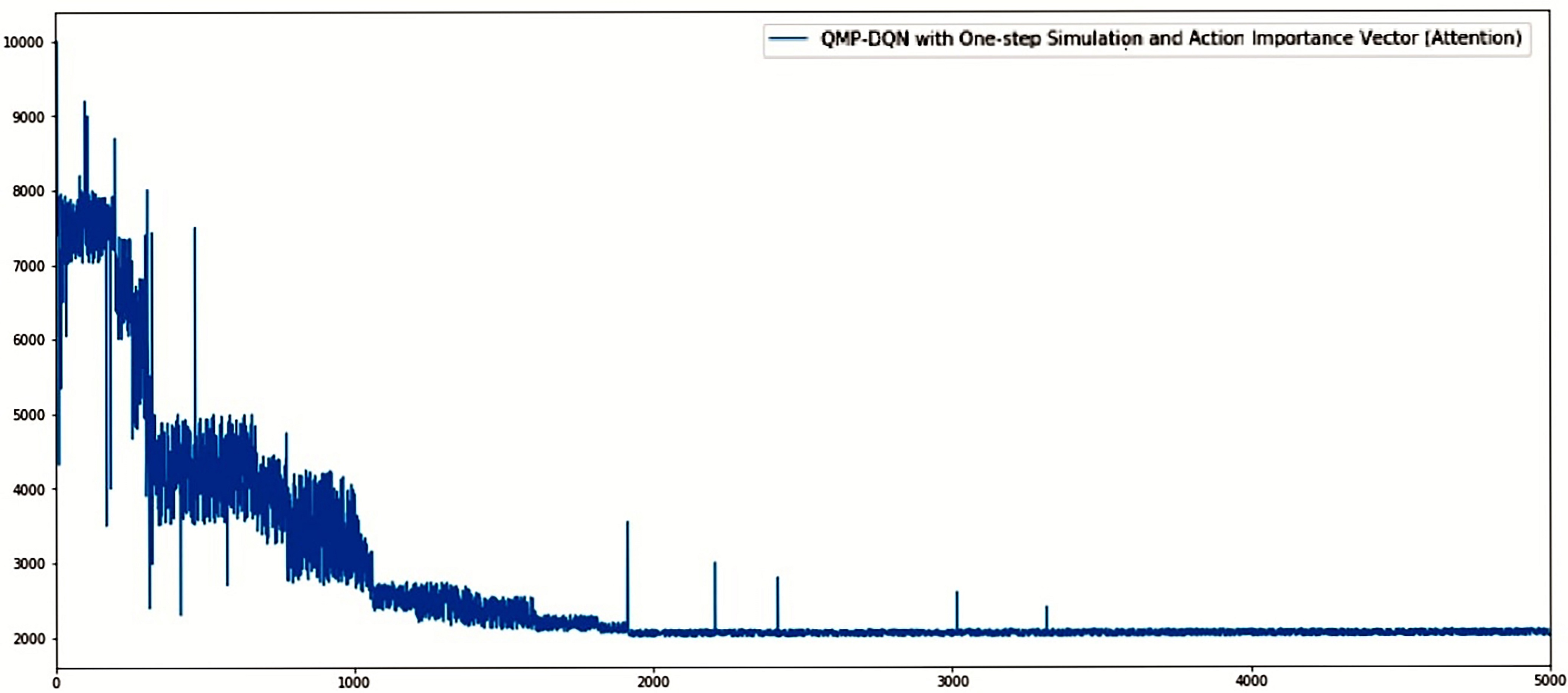}
	\label{fig:sfig1}
	\caption{Proposed method applied on the UiA Building}
\end{figure*}
The results of our proposed method consisting of shortest path Q-matrix transfer learning to Dueling-DQN model with one-step simulation and action importance vector acting as an attention mechanism applied on the University of Agder's A,B,C and D blocks consisting of $91$ rooms and $8281$ actions (whose graph is shown in Fig. 13) is shown in Fig. 14. The performance numbers are given below:\\
\begin{description}
	\item[$\bullet$ Average number of time-steps:]2234.5
	\item[$\bullet$ Minimum number of time-steps:]$\sim$2000
	\item[$\bullet$ Training time (per episode):]32.18 \emph{s}
\end{description}
The graph in Fig. 14 shows the convergence of our method with evacuation time-steps on the y-axis and the episode number on the x-axis. It takes slightly longer to converge compared to the convergence in previous small example environments. This is obviously due to the size of the environment and complex connectivity. But overall the performance of our model is excellent.\\
After $\sim1900$ episodes, the algorithm has almost converged. There are a few spikes suggesting fluctuations from the optimal behaviour due to the dynamic nature of the environment and the uncertainty in actions. After $\sim3300$ episodes, the algorithm completely converges in the range $(2000-2070)$ times-steps for total evacuation. The method cannot converge to the minimum possible time-steps $=2000$ because of the fire spread dynamics, encountering bottleneck conditions and action uncertainty.\\
The results clearly suggest that even though the proposed fire evacuation environment is dynamic, uncertain and full of constraints, our proposed method using novel action reduction technique with attention based mechanism and transfer learning of shortest path information is able to achieve excellent performance on a large and complex real world building model. This further confirms that, with a minute added overhead of one-step simulation and action importance vector, our method is scalable to much larger and complex building models.

\section{Conclusion}
In this paper, we propose the first realistic fire evacuation environment to train reinforcement learning agents. The environment is implemented in OpenAI gym format. The environment has been developed to simulate realistic fire scenarios. It includes features like fire spread with the help of exponential decay reward functions and degree functions, bottlenecks, uncertainty in performing an action and a graph based environment for accurately mapping a building model.\\
We also propose a new reinforcement learning method for training on our environment. We use tabular Q-learning to generate q-values for shortest path to the exit using the adjacency matrix of the graph based environment. Then, the result of Q-learning (after being offset by a $\sigma$) is used to pretrain the DQN network weights to incorporate shortest path information in the agent. Finally, the pretrained weights of the DQN based agents are trained on the fire evacuation environment.\\
We prove the faster convergence of our method using Task Transfer Q-learning theorems and the convergence of Q-learning for the shortest path task. The Q-matrix pretrained DQN agents (QMP-DQN) are compared with state-of-the-art reinforcement learning algorithms like DQN, DDQN, Dueling-DQN, PPO, VPG, A2C, ACKTR and SARSA on the fire evacuation environment. The proposed method is able to outperform all these models on our environment on the basis of convergence, training time and stability. Also, the comparisons of QMP-DQN with original DQN based models show clear improvements over the latter.\\
Finally, we show the scalability of our method by testing it on a real world large and complex building model. In order to reduce the large action space ($8281$ actions), we use the one-step simulation technique on the pretraining environment instance to calculate the action importance vector, which can be thought of as an attention based mechanism. The action importance vector gives the best $k$ actions a weight of $0$ and the rest are assigned a large negative weight of $-9999$ (to render the Q-values of these too low to be selected by the Q-function). This reduces the action space by $\sim90\%$ and our proposed method, QMP-DQN model, is applied on this reduced action space. We test this method on the UiA, Campus Grimstad building, with the environment consisting of $91$ rooms. The results show that this combination of methods works really well in a large real world fire evacuation emergency environment. 

\section*{Acknowledgment}
The authors would like to thank Tore Olsen, Chief of the Grimstad Fire Department, and the Grimstad Fire Department for supporting us with their expertise regarding fire emergencies and evacuation procedures as well as giving feedback to improve our proposed environment and evacuation system. We would also like to thank Dr. Jaziar Radianti, Center for Integrated Emergency Management (CIEM), University of Agder, for her input to this research work.

%





\ifCLASSOPTIONcaptionsoff
  \newpage
\fi



%

\bibliography{mybib}{}

\begin{thebibliography}{10}

\bibitem{BRIGADE}
Abbas Abdolmaleki, Mostafa Movahedi, Sajjad Salehi, Nuno Lau, and Luis~Paulo
  Reis.
\newblock A reinforcement learning based method for optimizing the process of
  decision making in fire brigade agents.
\newblock In Luis Antunes and H.~Sofia Pinto, editors, {\em Progress in
  Artificial Intelligence}, pages 340--351, Berlin, Heidelberg, 2011. Springer
  Berlin Heidelberg.

\bibitem{ROBO}
Fatemeh Pahlevan~Aghababa, Masaru Shimizu, Francesco Amigoni, Amirreza Kabiri,
  and Arnoud Visser.
\newblock Robocup 2018 robocup rescue simulation league virtual robot
  competition rules document, May 2018.

\bibitem{SARSA}
G.~A. Rummery and M.~Niranjan.
\newblock On-line {Q}-learning using connectionist systems.
\newblock Technical Report TR 166, Cambridge University Engineering Department,
  Cambridge, England, 1994.

\bibitem{AGENT}
Daniel Moura and Eugenio Oliveira.
\newblock Fighting fire with agents: An agent coordination model for simulated
  firefighting.
\newblock In {\em Proceedings of the 2007 Spring Simulation Multiconference -
  Volume 2}, SpringSim '07, pages 71--78, San Diego, CA, USA, 2007. Society for
  Computer Simulation International.

\bibitem{INDOOR}
Haifeng Zhao and Stephan Winter.
\newblock A time-aware routing map for indoor evacuation.
\newblock {\em Sensors}, 16(1), 2016.

\bibitem{THESIS}
Ashley Wharton.
\newblock Simulation and investigation of multi-agent reinforcement learning
  for building evacuation scenarios *.
\newblock 2009.

\bibitem{SAFEGRESS}
Mei~Ling Chu, Paolo Parigi, Kincho Law, and Jean-Claude Latombe.
\newblock Modeling social behaviors in an evacuation simulator.
\newblock {\em Computer Animation and Virtual Worlds}, 25(3-4):373--382.

\bibitem{EGRESS}
V.~J. Cassol, E.~Smania Testa, C.~Rosito Jung, M.~Usman, P.~Faloutsos,
  G.~Berseth, M.~Kapadia, N.~I. Badler, and S.~Raupp Musse.
\newblock Evaluating and optimizing evacuation plans for crowd egress.
\newblock {\em IEEE Computer Graphics and Applications}, 37(4):60--71, 2017.

\bibitem{BDI}
LH~Lee and Young-Jun Son.
\newblock Dynamic learning in human decision behavior for evacuation scenarios
  under bdi framework.
\newblock In {\em Proceedings of the 2009 INFORMS Simulation Society Research
  Workshop. INFORMS Simulation Society: Catonsville, MD}, pages 96--100, 2009.

\bibitem{BDI2}
Seungho Lee, Young-Jun Son, and Judy Jin.
\newblock An integrated human decision making model for evacuation scenarios
  under a bdi framework.
\newblock {\em ACM Trans. Model. Comput. Simul.}, 20(4):23:1--23:24, November
  2010.

\bibitem{GYM}
Greg Brockman, Vicki Cheung, Ludwig Pettersson, Jonas Schneider, John Schulman,
  Jie Tang, and Wojciech Zaremba.
\newblock Openai gym.
\newblock {\em CoRR}, abs/1606.01540, 2016.

\bibitem{UNITY}
Arthur Juliani, Vincent{-}Pierre Berges, Esh Vckay, Yuan Gao, Hunter Henry,
  Marwan Mattar, and Danny Lange.
\newblock Unity: {A} general platform for intelligent agents.
\newblock {\em CoRR}, abs/1809.02627, 2018.

\bibitem{SC2}
Oriol Vinyals, Timo Ewalds, Sergey Bartunov, Petko Georgiev, Alexander~Sasha
  Vezhnevets, Michelle Yeo, Alireza Makhzani, Heinrich K{\"{u}}ttler, John
  Agapiou, Julian Schrittwieser, John Quan, Stephen Gaffney, Stig Petersen,
  Karen Simonyan, Tom Schaul, Hado van Hasselt, David Silver, Timothy~P.
  Lillicrap, Kevin Calderone, Paul Keet, Anthony Brunasso, David Lawrence,
  Anders Ekermo, Jacob Repp, and Rodney Tsing.
\newblock Starcraft {II:} {A} new challenge for reinforcement learning.
\newblock {\em CoRR}, abs/1708.04782, 2017.

\bibitem{PYG}
Pete Shinners.
\newblock Pygame.
\newblock \url{http://pygame.org/}, 2011.

\bibitem{SIM}
{\L}ukasz Kidzi\'nski, Sharada~P Mohanty, Carmichael Ong, Jennifer Hicks, Sean
  Francis, Sergey Levine, Marcel Salath\'e, and Scott Delp.
\newblock Learning to run challenge: Synthesizing physiologically accurate
  motion using deep reinforcement learning.
\newblock In Sergio Escalera and Markus Weimer, editors, {\em NIPS 2017
  Competition Book}. Springer, Springer, 2018.

\bibitem{CELLULAR}
Stephen Wolfram.
\newblock Statistical mechanics of cellular automata.
\newblock {\em Rev. Mod. Phys.}, 55:601--644, Jul 1983.

\bibitem{CELLULAR2}
S.~I. Pak and T.~Hayakawa.
\newblock Forest fire modeling using cellular automata and percolation
  threshold analysis.
\newblock In {\em Proceedings of the 2011 American Control Conference}, pages
  293--298, June 2011.

\bibitem{CONTROL}
Marco Wiering and Marco Dorigo.
\newblock Learning to control forest fires.
\newblock {\em Ultrech University Repository}, Jan 1998.

\bibitem{AIFOREST}
G.~E. Sakr, I.~H. Elhajj, G.~Mitri, and U.~C. Wejinya.
\newblock Artificial intelligence for forest fire prediction.
\newblock In {\em 2010 IEEE/ASME International Conference on Advanced
  Intelligent Mechatronics}, pages 1311--1316, July 2010.

\bibitem{SPREAD}
Sriram Ganapathi~Subramanian and Mark Crowley.
\newblock Using spatial reinforcement learning to build forest wildfire
  dynamics models from satellite images.
\newblock {\em Frontiers in ICT}, 5:6, 2018.

\bibitem{TASK}
Amir~R. Zamir, Alexander Sax, William~B. Shen, Leonidas~J. Guibas, Jitendra
  Malik, and Silvio Savarese.
\newblock Taskonomy disentangling task transfer learning.
\newblock In {\em IEEE Conference on Computer Vision and Pattern Recognition
  (CVPR)}. IEEE, 2018.

\bibitem{LM}
Jeremy Howard and Sebastian Ruder.
\newblock Universal language model fine-tuning for text classification.
\newblock {\em CoRR}, abs/1801.06146, 2018.

\bibitem{BERT}
Jacob Devlin, Ming{-}Wei Chang, Kenton Lee, and Kristina Toutanova.
\newblock {BERT:} pre-training of deep bidirectional transformers for language
  understanding.
\newblock {\em CoRR}, abs/1810.04805, 2018.

\bibitem{WORLD}
David Ha and J{\"{u}}rgen Schmidhuber.
\newblock World models.
\newblock {\em CoRR}, abs/1803.10122, 2018.

\bibitem{TTQL}
Yue Wang, Qi~Meng, Wei Chen, Yuting Liu, Zhiming Ma, and Tie{-}Yan Liu.
\newblock Target transfer q-learning and its convergence analysis.
\newblock {\em CoRR}, abs/1809.08923, 2018.

\bibitem{SUTTON}
Richard~S. Sutton and Andrew~G. Barto.
\newblock {\em Introduction to Reinforcement Learning}.
\newblock MIT Press, Cambridge, MA, USA, 1st edition, 1998.

\bibitem{MDP}
Martin~L. Puterman.
\newblock {\em Markov Decision Processes: Discrete Stochastic Dynamic
  Programming}.
\newblock John Wiley \& Sons, Inc., New York, NY, USA, 1st edition, 1994.

\bibitem{Q}
Christopher~JCH Watkins and Peter Dayan.
\newblock Q-learning.
\newblock {\em Machine learning}, 8(3-4):279--292, 1992.

\bibitem{FIRST}
Leemon Baird.
\newblock Residual algorithms: Reinforcement learning with function
  approximation.
\newblock In {\em Machine Learning Proceedings 1995}, pages 30 -- 37. Morgan
  Kaufmann, San Francisco (CA), 1995.

\bibitem{IMAGENET}
Olga Russakovsky, Jia Deng, Hao Su, Jonathan Krause, Sanjeev Satheesh, Sean Ma,
  Zhiheng Huang, Andrej Karpathy, Aditya Khosla, Michael Bernstein,
  Alexander~C. Berg, and Li~Fei-Fei.
\newblock {ImageNet Large Scale Visual Recognition Challenge}.
\newblock {\em International Journal of Computer Vision (IJCV)},
  115(3):211--252, 2015.

\bibitem{CNNOriginal}
Kunihiko Fukushima.
\newblock Neocognitron: A self-organizing neural network model for a mechanism
  of pattern recognition unaffected by shift in position.
\newblock {\em Biological Cybernetics}, 36(4):193--202, 1980.

\bibitem{VGG16}
Karen Simonyan and Andrew Zisserman.
\newblock Very deep convolutional networks for large-scale image recognition.
\newblock {\em CoRR}, abs/1409.1556, 2014.

\bibitem{RESNET}
Kaiming He, Xiangyu Zhang, Shaoqing Ren, and Jian Sun.
\newblock Deep residual learning for image recognition.
\newblock In {\em The IEEE Conference on Computer Vision and Pattern
  Recognition (CVPR)}, June 2016.

\bibitem{LeNet-5}
Y.~Lecun, L.~Bottou, Y.~Bengio, and P.~Haffner.
\newblock Gradient-based learning applied to document recognition.
\newblock {\em Proceedings of the IEEE}, 86(11):2278--2324, Nov 1998.

\bibitem{XCEPTION}
Fran{\c{c}}ois Chollet.
\newblock Xception: Deep learning with depthwise separable convolutions.
\newblock {\em CoRR}, abs/1610.02357, 2016.

\bibitem{ALEXNET}
Alex Krizhevsky, Ilya Sutskever, and Geoffrey~E Hinton.
\newblock Imagenet classification with deep convolutional neural networks.
\newblock In F.~Pereira, C.~J.~C. Burges, L.~Bottou, and K.~Q. Weinberger,
  editors, {\em Advances in Neural Information Processing Systems 25}, pages
  1097--1105. Curran Associates, Inc., 2012.

\bibitem{LAS}
W.~Chan, N.~Jaitly, Q.~Le, and O.~Vinyals.
\newblock Listen, attend and spell: A neural network for large vocabulary
  conversational speech recognition.
\newblock In {\em 2016 IEEE International Conference on Acoustics, Speech and
  Signal Processing (ICASSP)}, pages 4960--4964, March 2016.

\bibitem{GOOGLESPEECH}
Chung-Cheng Chiu, Tara Sainath, Yonghui Wu, Rohit Prabhavalkar, Patrick Nguyen,
  Zhifeng Chen, Anjuli Kannan, Ron~J. Weiss, Kanishka Rao, Katya Gonina,
  Navdeep Jaitly, Bo~Li, Jan Chorowski, and Michiel Bacchiani.
\newblock State-of-the-art speech recognition with sequence-to-sequence models.
\newblock 2018.

\bibitem{E2ESRRNN}
Alex Graves and Navdeep Jaitly.
\newblock Towards end-to-end speech recognition with recurrent neural networks.
\newblock In Tony Jebara and Eric~P. Xing, editors, {\em Proceedings of the
  31st International Conference on Machine Learning (ICML-14)}, pages
  1764--1772. JMLR Workshop and Conference Proceedings, 2014.

\bibitem{GHINT}
G.~Hinton, L.~Deng, D.~Yu, G.~E. Dahl, A.~Mohamed, N.~Jaitly, A.~Senior,
  V.~Vanhoucke, P.~Nguyen, T.~N. Sainath, and B.~Kingsbury.
\newblock Deep neural networks for acoustic modeling in speech recognition: The
  shared views of four research groups.
\newblock {\em IEEE Signal Processing Magazine}, 29(6):82--97, Nov 2012.

\bibitem{ALEX}
Alex Graves, Abdel{-}rahman Mohamed, and Geoffrey~E. Hinton.
\newblock Speech recognition with deep recurrent neural networks.
\newblock {\em CoRR}, abs/1303.5778, 2013.

\bibitem{LVCSR}
T.~N. Sainath, A.~Mohamed, B.~Kingsbury, and B.~Ramabhadran.
\newblock Deep convolutional neural networks for lvcsr.
\newblock In {\em 2013 IEEE International Conference on Acoustics, Speech and
  Signal Processing}, pages 8614--8618, May 2013.

\bibitem{CONTEXT}
G.~E. Dahl, D.~Yu, L.~Deng, and A.~Acero.
\newblock Context-dependent pre-trained deep neural networks for
  large-vocabulary speech recognition.
\newblock {\em IEEE Transactions on Audio, Speech, and Language Processing},
  20(1):30--42, Jan 2012.

\bibitem{NLP1}
Xiang Zhang, Junbo Zhao, and Yann LeCun.
\newblock Character-level convolutional networks for text classification.
\newblock In {\em Proceedings of the 28th International Conference on Neural
  Information Processing Systems - Volume 1}, NIPS'15, pages 649--657,
  Cambridge, MA, USA, 2015. MIT Press.

\bibitem{NLP2}
Rohan Anil, Gabriel Pereyra, Alexandre Passos, Robert Orm{\'{a}}ndi, George~E.
  Dahl, and Geoffrey~E. Hinton.
\newblock Large scale distributed neural network training through online
  distillation.
\newblock {\em CoRR}, abs/1804.03235, 2018.

\bibitem{NLP3}
Mia~Xu Chen, Orhan Firat, Ankur Bapna, Melvin Johnson, Wolfgang Macherey,
  George Foster, Llion Jones, Niki Parmar, Mike Schuster, Zhifeng Chen, Yonghui
  Wu, and Macduff Hughes.
\newblock The best of both worlds: Combining recent advances in neural machine
  translation.
\newblock {\em CoRR}, abs/1804.09849, 2018.

\bibitem{NLP4}
Lierni Sestorain, Massimiliano Ciaramita, Christian Buck, and Thomas Hofmann.
\newblock Zero-shot dual machine translation.
\newblock {\em CoRR}, abs/1805.10338, 2018.

\bibitem{NLP5}
Hany Hassan, Anthony Aue, Chang Chen, Vishal Chowdhary, Jonathan Clark,
  Christian Federmann, Xuedong Huang, Marcin Junczys{-}Dowmunt, William Lewis,
  Mu~Li, Shujie Liu, Tie{-}Yan Liu, Renqian Luo, Arul Menezes, Tao Qin, Frank
  Seide, Xu~Tan, Fei Tian, Lijun Wu, Shuangzhi Wu, Yingce Xia, Dongdong Zhang,
  Zhirui Zhang, and Ming Zhou.
\newblock Achieving human parity on automatic chinese to english news
  translation.
\newblock {\em CoRR}, abs/1803.05567, 2018.

\bibitem{DNNRL1}
S.~Lange and M.~Riedmiller.
\newblock Deep auto-encoder neural networks in reinforcement learning.
\newblock In {\em The 2010 International Joint Conference on Neural Networks
  (IJCNN)}, pages 1--8, July 2010.

\bibitem{DNNRL2}
H.~D. Patino and D.~Liu.
\newblock Neural network-based model reference adaptive control system.
\newblock {\em IEEE Transactions on Systems, Man, and Cybernetics, Part B
  (Cybernetics)}, 30(1):198--204, Feb 2000.

\bibitem{DQN1}
Volodymyr Mnih, Koray Kavukcuoglu, David Silver, Alex Graves, Ioannis
  Antonoglou, Daan Wierstra, and Martin Riedmiller.
\newblock Playing atari with deep reinforcement learning, 2013.
\newblock cite arxiv:1312.5602Comment: NIPS Deep Learning Workshop 2013.

\bibitem{DQN2}
Volodymyr Mnih, Koray Kavukcuoglu, David Silver, Andrei~A. Rusu, Joel Veness,
  Marc~G. Bellemare, Alex Graves, Martin Riedmiller, Andreas~K. Fidjeland,
  Georg Ostrovski, Stig Petersen, Charles Beattie, Amir Sadik, Ioannis
  Antonoglou, Helen King, Dharshan Kumaran, Daan Wierstra, Shane Legg, and
  Demis Hassabis.
\newblock Human-level control through deep reinforcement learning.
\newblock {\em Nature}, 518(7540):529--533, feb 2015.

\bibitem{SGD}
Herbert Robbins and Sutton Monro.
\newblock A stochastic approximation method.
\newblock {\em Ann. Math. Statist.}, 22(3):400--407, 09 1951.

\bibitem{RMS}
T.~Tieleman and G.~Hinton.
\newblock {Lecture 6.5---RmsProp: Divide the gradient by a running average of
  its recent magnitude}.
\newblock COURSERA: Neural Networks for Machine Learning, 2012.

\bibitem{ADAGRAD}
John Duchi, Elad Hazan, and Yoram Singer.
\newblock Adaptive subgradient methods for online learning and stochastic
  optimization.
\newblock Technical Report UCB/EECS-2010-24, EECS Department, University of
  California, Berkeley, Mar 2010.

\bibitem{ADAM}
D.P. Kingma and L.J. Ba.
\newblock Adam: A method for stochastic optimization.
\newblock In {\em ICLR}, International Conference on Learning Representations
  (ICLR), page~13, San Diego, CA, USA, 7--9 May 2015. Ithaca, NY: arXiv.org.

\bibitem{RLNN}
Long-Ji Lin.
\newblock {\em Reinforcement Learning for Robots Using Neural Networks}.
\newblock PhD thesis, Pittsburgh, PA, USA, 1992.
\newblock UMI Order No. GAX93-22750.

\bibitem{DDQN1}
Hado~V. Hasselt.
\newblock Double q-learning.
\newblock In J.~D. Lafferty, C.~K.~I. Williams, J.~Shawe-Taylor, R.~S. Zemel,
  and A.~Culotta, editors, {\em Advances in Neural Information Processing
  Systems 23}, pages 2613--2621. Curran Associates, Inc., 2010.

\bibitem{DDQN2}
Hado~van Hasselt, Arthur Guez, and David Silver.
\newblock Deep reinforcement learning with double q-learning.
\newblock In {\em Proceedings of the Thirtieth AAAI Conference on Artificial
  Intelligence}, AAAI'16, pages 2094--2100. AAAI Press, 2016.

\bibitem{DUELING}
Ziyu Wang, Tom Schaul, Matteo Hessel, Hado Van~Hasselt, Marc Lanctot, and Nando
  De~Freitas.
\newblock Dueling network architectures for deep reinforcement learning.
\newblock In {\em Proceedings of the 33rd International Conference on
  International Conference on Machine Learning - Volume 48}, ICML'16, pages
  1995--2003. JMLR.org, 2016.

\bibitem{CURRICULUM}
Yoshua Bengio, J{\'e}r\^{o}me Louradour, Ronan Collobert, and Jason Weston.
\newblock Curriculum learning.
\newblock In {\em Proceedings of the 26th Annual International Conference on
  Machine Learning}, ICML '09, pages 41--48, New York, NY, USA, 2009. ACM.

\bibitem{CONVERGEQ}
Csaba Szepesv{\'a}ri.
\newblock The asymptotic convergence-rate of q-learning.
\newblock In {\em Advances in Neural Information Processing Systems}, pages
  1064--1070, 1998.

\bibitem{STOCONVERGE}
Tommi Jaakkola, Michael~I. Jordan, and Satinder~P. Singh.
\newblock On the convergence of stochastic iterative dynamic programming
  algorithms.
\newblock {\em Neural Comput.}, 6(6):1185--1201, November 1994.

\bibitem{ASYNCONVERGE}
John~N. Tsitsiklis.
\newblock Asynchronous stochastic approximation and q-learning.
\newblock {\em Machine Learning}, 16(3):185--202, Sep 1994.

\bibitem{PRETRAIN}
E.~Larsson.
\newblock Evaluation of pretraining methods for deep reinforcement learning.
\newblock 2018.

\bibitem{DQfD1}
Todd Hester, Matej Vecerik, Olivier Pietquin, Marc Lanctot, Tom Schaul, Bilal
  Piot, Andrew Sendonaris, Gabriel Dulac-Arnold, Ian Osband, John Agapiou,
  Joel~Z. Leibo, and Audrunas Gruslys.
\newblock Learning from demonstrations for real world reinforcement learning.
\newblock {\em CoRR}, abs/1704.03732, 2017.

\bibitem{DQfD2}
Todd Hester, Matej Vecerik, Olivier Pietquin, Marc Lanctot, Tom Schaul, Bilal
  Piot, Andrew Sendonaris, Gabriel Dulac{-}Arnold, Ian Osband, John Agapiou,
  Joel~Z. Leibo, and Audrunas Gruslys.
\newblock Learning from demonstrations for real world reinforcement learning.
\newblock {\em CoRR}, abs/1704.03732, 2017.

\bibitem{RELU}
Xavier Glorot, Antoine Bordes, and Yoshua Bengio.
\newblock Deep sparse rectifier neural networks.
\newblock In Geoffrey Gordon, David Dunson, and Miroslav Dudik, editors, {\em
  Proceedings of the Fourteenth International Conference on Artificial
  Intelligence and Statistics}, volume~15 of {\em Proceedings of Machine
  Learning Research}, pages 315--323, Fort Lauderdale, FL, USA, 11--13 Apr
  2011. PMLR.

\bibitem{TENSOR}
Mart\'{\i}n Abadi, Ashish Agarwal, Paul Barham, Eugene Brevdo, Zhifeng Chen,
  Craig Citro, Greg~S. Corrado, Andy Davis, Jeffrey Dean, Matthieu Devin,
  Sanjay Ghemawat, Ian Goodfellow, Andrew Harp, Geoffrey Irving, Michael Isard,
  Yangqing Jia, Rafal Jozefowicz, Lukasz Kaiser, Manjunath Kudlur, Josh
  Levenberg, Dan Man\'{e}, Rajat Monga, Sherry Moore, Derek Murray, Chris Olah,
  Mike Schuster, Jonathon Shlens, Benoit Steiner, Ilya Sutskever, Kunal Talwar,
  Paul Tucker, Vincent Vanhoucke, Vijay Vasudevan, Fernanda Vi\'{e}gas, Oriol
  Vinyals, Pete Warden, Martin Wattenberg, Martin Wicke, Yuan Yu, and Xiaoqiang
  Zheng.
\newblock {TensorFlow}: Large-scale machine learning on heterogeneous systems,
  2015.
\newblock Software available from tensorflow.org.

\bibitem{KERAS}
Fran\c{c}ois Chollet et~al.
\newblock Keras.
\newblock \url{https://keras.io}, 2015.

\bibitem{PPO}
John Schulman, Filip Wolski, Prafulla Dhariwal, Alec Radford, and Oleg Klimov.
\newblock Proximal policy optimization algorithms.
\newblock {\em CoRR}, abs/1707.06347, 2017.

\bibitem{VPG}
Richard~S. Sutton, David McAllester, Satinder Singh, and Yishay Mansour.
\newblock Policy gradient methods for reinforcement learning with function
  approximation.
\newblock In {\em Proceedings of the 12th International Conference on Neural
  Information Processing Systems}, NIPS'99, pages 1057--1063, Cambridge, MA,
  USA, 1999. MIT Press.

\bibitem{A2C}
Volodymyr Mnih, Adri\`{a}~Puigdom\`{e}nech Badia, Mehdi Mirza, Alex Graves, Tim
  Harley, Timothy~P. Lillicrap, David Silver, and Koray Kavukcuoglu.
\newblock Asynchronous methods for deep reinforcement learning.
\newblock In {\em Proceedings of the 33rd International Conference on
  International Conference on Machine Learning - Volume 48}, ICML'16, pages
  1928--1937. JMLR.org, 2016.

\bibitem{A3C}
Volodymyr Mnih, Adri{\`{a}}~Puigdom{\`{e}}nech Badia, Mehdi Mirza, Alex Graves,
  Timothy~P. Lillicrap, Tim Harley, David Silver, and Koray Kavukcuoglu.
\newblock Asynchronous methods for deep reinforcement learning.
\newblock {\em CoRR}, abs/1602.01783, 2016.

\bibitem{NOA3C}
Yuhuai Wu, Elman Mansimov, Shun Liao, Alec Radford, and John Schulman.
\newblock Openai baselines: Acktr and a2c.
\newblock https://openai.com/blog/baselines-acktr-a2c/, 2017.

\bibitem{ACKTR}
Yuhuai Wu, Elman Mansimov, Shun Liao, Roger~B. Grosse, and Jimmy Ba.
\newblock Scalable trust-region method for deep reinforcement learning using
  kronecker-factored approximation.
\newblock {\em CoRR}, abs/1708.05144, 2017.

\bibitem{KFAC}
James Martens and Roger~B. Grosse.
\newblock Optimizing neural networks with kronecker-factored approximate
  curvature.
\newblock {\em CoRR}, abs/1503.05671, 2015.

\bibitem{ALPHAGO}
David Silver, Aja Huang, Christopher~J. Maddison, Arthur Guez, Laurent Sifre,
  George van~den Driessche, Julian Schrittwieser, Ioannis Antonoglou, Veda
  Panneershelvam, Marc Lanctot, Sander Dieleman, Dominik Grewe, John Nham, Nal
  Kalchbrenner, Ilya Sutskever, Timothy Lillicrap, Madeleine Leach, Koray
  Kavukcuoglu, Thore Graepel, and Demis Hassabis.
\newblock Mastering the game of go with deep neural networks and tree search.
\newblock {\em Nature}, 529:484--503, 2016.

\bibitem{ALPHAGOZERO}
David Silver, Julian Schrittwieser, Karen Simonyan, Ioannis Antonoglou, Aja
  Huang, Arthur Guez, Thomas Hubert, Lucas Baker, Matthew Lai, Adrian Bolton,
  et~al.
\newblock Mastering the game of go without human knowledge.
\newblock {\em Nature}, 550(7676):354, 2017.

\bibitem{LDAS}
Gabriel Dulac{-}Arnold, Richard Evans, Peter Sunehag, and Ben Coppin.
\newblock Deep reinforcement learning in large discrete action spaces.
\newblock {\em CoRR}, abs/1512.07679, 2015.

\bibitem{ATTEND1}
Ashish Vaswani, Noam Shazeer, Niki Parmar, Jakob Uszkoreit, Llion Jones,
  Aidan~N. Gomez, Lukasz Kaiser, and Illia Polosukhin.
\newblock Attention is all you need.
\newblock {\em CoRR}, abs/1706.03762, 2017.

\bibitem{ATTEND2}
Kelvin Xu, Jimmy Ba, Ryan Kiros, Kyunghyun Cho, Aaron~C. Courville, Ruslan
  Salakhutdinov, Richard~S. Zemel, and Yoshua Bengio.
\newblock Show, attend and tell: Neural image caption generation with visual
  attention.
\newblock {\em CoRR}, abs/1502.03044, 2015.

\bibitem{ATTEND3}
Dzmitry Bahdanau, Kyunghyun Cho, and Yoshua Bengio.
\newblock Neural machine translation by jointly learning to align and
  translate, 2014.
\newblock cite arxiv:1409.0473Comment: Accepted at ICLR 2015 as oral
  presentation.

\end{thebibliography}
\bibliographystyle{unsrt}










\end{document}